\pdfoutput=1
\documentclass[11pt]{article}

\usepackage{acl}

\usepackage{amsmath,amsfonts,bm}

\def\eqref#1{equation~\ref{#1}}

\def\1{\bm{1}}

\DeclareMathAlphabet{\mathsfit}{\encodingdefault}{\sfdefault}{m}{sl}
\SetMathAlphabet{\mathsfit}{bold}{\encodingdefault}{\sfdefault}{bx}{n}

\usepackage{times}
\usepackage{latexsym}
\usepackage[T1]{fontenc}
\usepackage[utf8]{inputenc}
\usepackage{microtype}
\usepackage{inconsolata}
\usepackage{graphicx}

\usepackage{hyperref}
\usepackage{url}
\usepackage{graphicx}
\usepackage{enumitem}
\usepackage{wrapfig}
\usepackage{authblk}

\definecolor{lightcoral}{rgb}{0.94, 0.5, 0.5}
\definecolor{lightgreen}{rgb}{0.56, 0.93, 0.56}
\definecolor{harvestgold}{rgb}{0.98, 0.85, 0.40}
\definecolor{brightlavender}{rgb}{0.75, 0.58, 0.89}
\definecolor{capri}{rgb}{0.0, 0.75, 1.0}
\definecolor{carminepink}{rgb}{0.92, 0.3, 0.26}
\definecolor{celadon}{rgb}{0.67, 0.88, 0.69}
\definecolor{darkpastelgreen}{rgb}{0.01, 0.75, 0.24}

\definecolor{hidden-draw}{RGB}{205, 44, 36}
\definecolor{hidden-blue}{RGB}{194,232,247}
\definecolor{hidden-orange}{RGB}{243,202,120}
\definecolor{hidden-yellow}{RGB}{242,244,193}
\definecolor{tree-level-1}{RGB}{245,20,85}
\definecolor{tree-level-2}{RGB}{246,86,118}
\definecolor{tree-level-3}{RGB}{248,177,193}
\definecolor{tree-leaf}{RGB}{176,230,198}

\definecolor{Self}{RGB}{255,0,128}
\definecolor{Ensemble}{RGB}{0,127,255}
\definecolor{Iterative}{RGB}{153,51,255}

\definecolor{exemplar1}{RGB}{136,98,148}
\definecolor{exemplar2}{RGB}{148,210,242}
\definecolor{knowledge1}{RGB}{249,219,152}
\definecolor{knowledge2}{RGB}{255,245,220}

\usepackage{nicematrix}
\usepackage{makecell}
\usepackage{xspace,mfirstuc,tabulary}
\usepackage{booktabs}
\usepackage{amssymb}
\usepackage{bbding}
\usepackage{xcolor,colortbl}
\usepackage{multirow}
\usepackage{inconsolata}
\usepackage{pifont}
\newrobustcmd{\B}{\bfseries}
\newcommand*\colourcheck[1]{%
  \expandafter\newcommand\csname #1check\endcsname{\textcolor{#1}{\ding{52}}}%
}
\newcommand*\colourcross[1]{%
  \expandafter\newcommand\csname #1cross\endcsname{\textcolor{#1}{\ding{55}}}%
}
\colourcheck{green}
\colourcross{red}

\usepackage[most]{tcolorbox}
\usepackage{booktabs}
\usepackage{multirow}
\usepackage{soul}
\usepackage{makecell}

\definecolor{color_1}{HTML}{A1F6EC}
\definecolor{color_2}{HTML}{f3b78b}
\definecolor{color_3}{HTML}{ffff9b}
\definecolor{color_4}{HTML}{cbcbf3}
\definecolor{color_5}{HTML}{92bfdb}
\definecolor{color_6}{HTML}{ff7e79}
\newcommand{\fa}{\sethlcolor{color_1}{\small \hl{~Fabrication~}}}
\newcommand{\fin}{\sethlcolor{color_2}{\small\hl{~Factual Inconsistency~}}}
\newcommand{\ci}{\sethlcolor{color_3}{\small\hl{~Context Inconsistency~}}}
\newcommand{\li}{\sethlcolor{color_5}{\small\hl{~Logical Inconsistency~}}}
\newcommand{\ii}{\sethlcolor{color_4}{\small\hl{~Instruction Inconsistency~}}}
\newcommand{\ler}{\sethlcolor{color_6}{\small\hl{~Calculation Error~}}}
\newcommand{\FA}{\sethlcolor{color_1}{\hl{~FA~}}}
\newcommand{\FI}{\sethlcolor{color_2}{\hl{~FI~}}}
\newcommand{\CI}{\sethlcolor{color_3}{\hl{~CI~}}}
\newcommand{\LI}{\sethlcolor{color_5}{\hl{~LI~}}}
\newcommand{\II}{\sethlcolor{color_4}{\hl{~II~}}}
\newcommand{\LE}{\sethlcolor{color_6}{\hl{~CE~}}}

\title{FG-PRM: Fine-grained Hallucination Detection and Mitigation in Language Model  Mathematical Reasoning}

\author{
\textbf{Ruosen Li}$^{1}$\textsuperscript{*}, \textbf{Ziming Luo}$^{2}$\textsuperscript{*}, \textbf{Xinya Du}$^{1}$\\
$^{1}${Department of Computer Science, University of Texas at Dallas}\\
$^{2}${Department of Electrical Engineering Computer Science, University of Michigen}\\
\texttt{\{ruosen.li, xinya.du\}@utdallas.edu} \quad
\texttt{luozm@umich.edu}
}

\begin{document}

\maketitle
\footnotetext{\textsuperscript{*}Both authors contributed equally to this work.}


\begin{abstract}


Hallucinations in large language models (LLMs) pose significant challenges in tasks requiring complex multi-step reasoning, such as mathematical problem-solving. Existing approaches primarily detect the presence of hallucinations but lack a nuanced understanding of their types and manifestations.
In this paper, we first introduce a comprehensive taxonomy that categorizes the common hallucinations in mathematical reasoning tasks into six types. We then propose FG-PRM (\textbf{F}ine-\textbf{G}rained \textbf{P}rocess \textbf{R}eward \textbf{M}odel), an augmented model designed to detect and mitigate hallucinations in a fine-grained, step-level manner.
To address the limitations of manually labeling training data, we propose an automated method for generating fine-grained hallucination data using LLMs.
Our FG-PRM demonstrates superior performance across two key tasks:
1) Fine-grained hallucination detection: classifying hallucination types for each reasoning step; and
2) Verification: ranking multiple LLM-generated outputs to select the most accurate solution.
Our experiments show that FG-PRM excels in fine-grained hallucination detection and substantially boosts the performance of LLMs on GSM8K and MATH benchmarks.
These results highlight the benefits of fine-grained supervision in enhancing
the reliability and interpretability of LLM reasoning processes.
Codes and datasets are available at: \url{https://github.com/du-nlp-lab/FG-PRM}.

\end{abstract}

\section{Introduction}

While considerable progress has been made in enhancing the general capabilities of large language models (LLMs), solving complex reasoning tasks such as answering mathematical questions remains a challenge. 
Recently, advanced prompting techniques~\citep{wei2022chain,yao2024tree,hao2023reasoning} are proposed to guide LLMs in breaking down complex reasoning tasks into simple steps, thus improving their performance and the interpretability of the reasoning process. Nevertheless, LLMs often produce incorrect or unverifiable statements—commonly known as hallucinations—that hinder their ability to solve complex problems that require multiple reasoning steps.

\begin{figure*}[htbp]
\centering
\vspace{-5mm}
\includegraphics[width=1\textwidth]{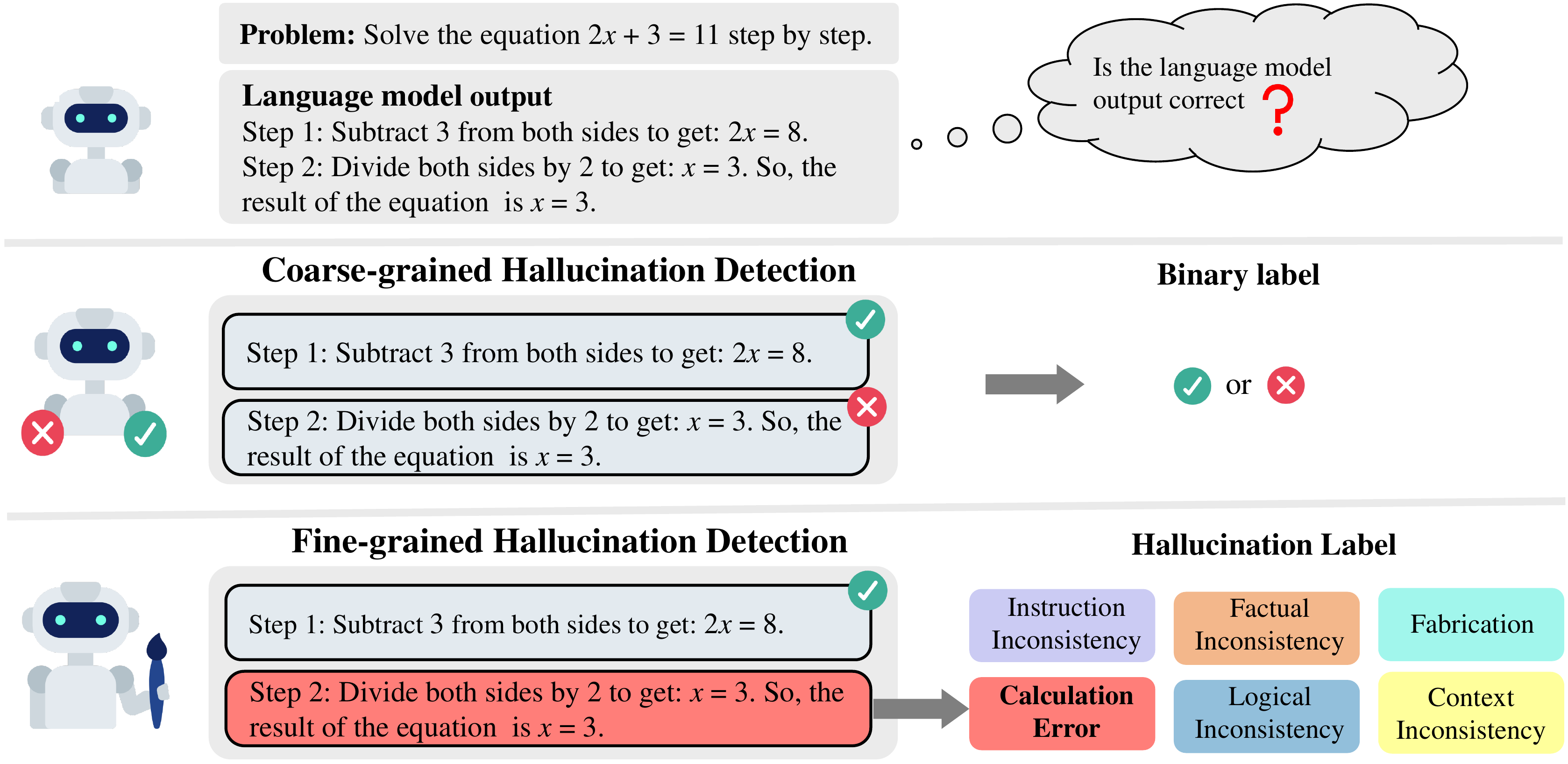}
\vspace{-7mm}
\caption{Overview of fine-grained hallucination detection for language model reasoning process. Above is an example for \ler\ hallucination.}
 \label{fig:diagram}
\vspace{-5mm}
\end{figure*}

Prior methods of mitigating hallucinations in reasoning chains largely focus on detecting their presence, with limited exploration into the distinct types of hallucinations produced. Our research goes beyond this by developing a fine-grained taxonomy that categorizes hallucinations based on their nature and manifestation (see Figure \ref{fig:diagram} for an illustration comparing coarse-grained detection with our method). We analyze reasoning steps to pinpoint the emergence of hallucinations and uncover patterns in their behavior. 

Training reward models is an effective approach for detecting and mitigating hallucinations, with the two primary categories being Outcome Reward Model (ORM)~\citep{cobbe2021training} and Process Reward Model (PRM)~\citep{lightman2023let}. ORMs evaluate the correctness of entire reasoning chains, while PRMs assess each step. PRMs have demonstrated superior performance in many scenarios~\citep{wang2023math} since they can provide more granular feedback and effectively guide models' reasoning process. However, collecting data to train PRMs is labor-intensive, particularly for  multi-step reasoning tasks, where human annotation is costly and prone to bias. To address this, we develop a novel method to automatically generate fine-grained hallucination data using LLMs. Specifically, giving a problem with a ground-truth solution, we first identify reasoning steps suitable for hallucination injection. Next, we utilize an LLM to generate additional reasoning steps incorporating various hallucination types based on our tailored instructions and demonstrations. The generated hallucinatory steps then serve as negative examples to train task-specific PRMs, each designed to detect a particular hallucination type. 


We evaluate our FG-PRM on two widely used mathematical benchmarks, GSM8K~\citep{cobbe2021training} and MATH~\citep{hendrycks2021measuring}. We validate the effectiveness of our method by two tasks: 
1) fine-grained hallucination detection, where we classify different hallucination types at each reasoning step; and
2) verification, where we rank multiple outputs generated by LLMs to select the most accurate solution.
Our major contributions are as follows:
\begin{itemize}[leftmargin=*, topsep=0mm, itemsep=1mm, parsep=0mm]
    \item We introduce a comprehensive hallucination taxonomy that categorizes common errors in mathematical reasoning tasks into six distinct types.
    
    \item We propose an automated method for synthesizing fine-grained hallucination data across without requiring human annotations. Based on this, we design FG-PRM to detect and mitigate hallucinations in a fine-grained, step-level manner.
    
    \item Extensive experiments show that FG-PRM surpasses ChatGPT-3.5 and Claude-3 in the hallucination detection task for most hallucination types, achieving over 5\% higher F1 scores.
    Moreover, FG-PRM outperforms PRMs by over 3\% in the verification task on GSM8K and MATH, surpassing numerous verifiers trained on human-labeled or coarse-grained data.
\end{itemize}



\section{Fine-grained Hallucination Taxonomy} 
\label{sec:taxonomy}

Large language models excel at solving tasks that require complex multi-step reasoning by generating solutions in a step-by-step and chain-of-thought format. Nevertheless, even state-of-the-art models are prone to inaccuracies, often producing content that is unfaithful, fabricated, inconsistent, or nonsensical. 
Categorizing and localizing these inaccuracies in reasoning steps is challenging but provides explicit insights into which parts of the model output have specific types of problems.

Building upon the prior work \citet{ji2023survey}, we develop a fine-grained taxonomy for two major categories of hallucinations: intrinsic and extrinsic hallucination, according to whether the hallucination can be verified by the input information or the contents LLMs have previously generated. To describe more complex errors surfacing in LM reasoning, we further divide the intrinsic hallucination into contextual inconsistency, logical inconsistency and instruction inconsistency, while extrinsic hallucinations are divided into calculation error, factual inconsistency, and fabrication.
We performed a pilot annotation with five NLP experts who have published at least three papers in related fields to refine our taxonomy, ensuring comprehensive coverage of various hallucination types.
The definitions of our proposed categories are elaborated below:

\begin{enumerate}[leftmargin=7mm, topsep=1mm, itemsep=0mm, parsep=0mm]
    \item [\textbf{(1)}] \textbf{\ci} refers to instances where a reasoning step is inconsistent with the contextual information provided by the user.

    \item [\textbf{(2)}] \textbf{\li} refers to the logical contradictions or inconsistencies between the current and previous reasoning steps.

    \item [\textbf{(3)}] \textbf{\ii} refers to instances where a reasoning step does not align with the explicit instructions of the user.
    
    \item [\textbf{(4)}] \textbf{\ler} refers to instances where a reasoning step makes incorrect calculations, which should be verifiable by external information or tools.
    
    \item [\textbf{(5)}] \textbf{\fin} refers to instances where a reasoning step contains facts that can be grounded in real-world information but present contradictions.
    
    \item [\textbf{(6)}] \textbf{\fa} refers to instances where a reasoning step contains facts that are unverifiable against knowledge in the real world or context.

\end{enumerate}



To illustrate our taxonomy more intuitively, we provide examples for each type of hallucination in Appendix Table~\ref{tab:hallucination_example}, along with corresponding explanations. 
Compared to the simplified taxonomy proposed in previous work~\citep{golovneva2022roscoe, prasad2023receval}, our refined taxonomy comprehensively captures the unique complexities of LLM hallucinations, offering a structured framework to study distinct patterns and enabling more granular analyses and targeted mitigation strategies.

\section{Methodology}

In this section, we first introduce two basic types of reward models (Section \ref{sec:rm_prelim}), the Outcome Reward Model (ORM) and the Process Reward Model (PRM). After that, we describe our automated framework for generating hallucination-annotated datasets, followed by a detailed explanation of the training procedure for our Fine-Grained Process Reward Model (FG-PRM) (Section \ref{FGPRM}).

\subsection{Preliminary}
\label{sec:rm_prelim}

\paragraph{ORM} The ORM was introduced by \citet{cobbe2021training}. Given a question \(x\) and its solution \(y\), an ORM assigns a sigmoid score \(r_y\) to the entire solution, indicating whether \(y\) is correct. ORMs are typically trained with cross-entropy loss over the entire solution. Assume $y^*$ is the ground-truth label of $y$, $y^* = 1$ if $y$ is correct, otherwise $y^* = 0$. The training objective minimizes the cross-entropy between the predicted outcome $r_y$ and the ground-truth $y^*$:
\begin{align} \mathcal{L}_{\text{ORM}} = y^*\log r_y + (1-y^*)\log(1-r_y) \end{align}

However, ORM's coarse-grained feedback limits its ability to diagnose errors within individual reasoning steps, as it only evaluates the final solution without considering intermediate correctness.


\paragraph{PRM} The PRM was introduced by \citet{lightman2023let}, addresses the limitations of ORM by providing step-level feedback. Instead of assigning a single score to the entire solution, PRM assigns a sigmoid score $r_{y_i}$ for each reasoning step $y_i$ in the solution $y$. This approach enables the model to evaluate the correctness of each intermediate step, providing more detailed feedback on where the reasoning process succeeds or fails. The training objective for PRM minimizes the sum of cross-entropy losses over all reasoning steps, allowing the model to learn from fine-grained supervision:
\begin{align} 
\mathcal{L}_{\text{PRM}} = \sum_{i=1}^{L} \log y^*_i \log r_{y_i} + (1-y^*_i)\log (1- r_{y_i})
\label{equa： PRM loss}
\end{align}

where $L$ is the number of reasoning steps in the solution $y$ and $y^*_i$ is the ground-truth label of the $i$-th step of $y$. 
By providing feedback at the step level, PRM offers significant advantages over ORM in tasks requiring complex, multi-step reasoning. 

\subsection{FG-PRM: Fine-grained Process Reward Model}
\label{FGPRM}
In this Section, we introduce our FG-PRM, the \textbf{F}ine-\textbf{G}rained \textbf{P}rocess \textbf{R}eward \textbf{M}odel for hallucination detection and mitigation. To reduce the annotation cost issues associated with PRM, we first introduce an automated process annotation framework for step-level fine-grained dataset synthesis. After that, we provide the training details for our FG-PRM on the synthetic dataset.

\begin{figure*}
    \vspace{-5mm}
    \centering
    \includegraphics[width=1\linewidth]{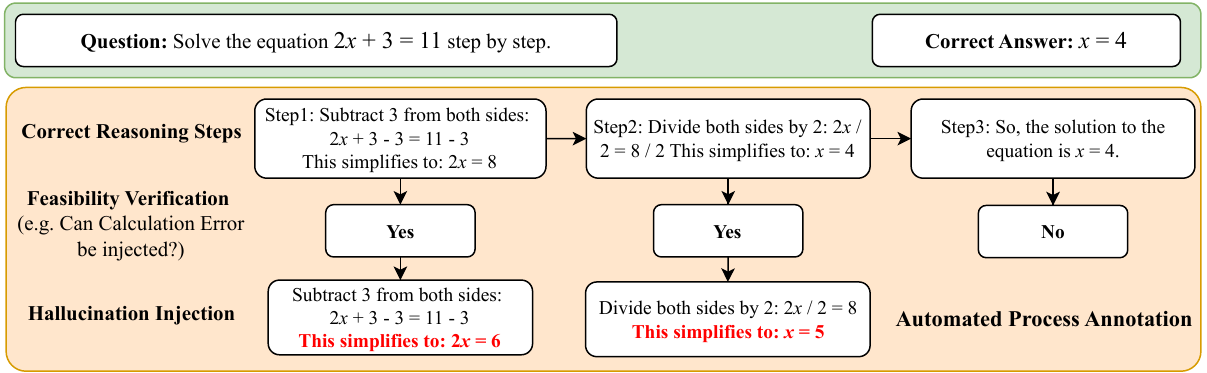}
    \vspace{-7mm}
    \caption{
     Our automated reasoning process annotation framework involves two steps: First, for each step, we instruct a language model to verify the feasibility of injecting hallucinations (using \ler\ as an example in this figure). Second, for identified steps where hallucinations can be injected, we prompt the language model to introduce hallucinations by providing instructions and few-shot demonstrations (see in Appendixes \ref{Synthesis prompt} and \ref{Hallucination Demonstration}).
    }
    \label{fig:pipeline}
    \vspace{-5mm}
\end{figure*}
\vspace{-2mm}

\subsubsection{Automated Hallucination Generation}
\label{sec:auto-hallu-gen}




Existing step-level datasets with fine-grained annotations~\citep{golovneva2022roscoe} are limited in size, and collecting the necessary data for training models with such detailed labels is costly, as it requires human annotators to provide fine-grained feedback for each reasoning step. To overcome the scarcity of human-labeled data, we introduce an automated hallucination annotation framework, as illustrated in Figure~\ref{fig:pipeline}. 
We start with the mathematical problems with golden chain-of-thought (CoT) solution dataset. To synthesize the negative reasoning steps, we adopt a two-step process as follows.


\paragraph{Step 1: Identify target reasoning steps}
In our taxonomy, each hallucination type follows a distinct pattern, requiring specific conditions met by the golden reasoning steps for generation. However, not all golden steps  can induce the generation of every type of hallucination. For instance, when a reasoning step is exclusively focused on numerical calculations, it becomes challenging to insert factual inconsistency.
To effectively introduce different hallucination into the reasoning process, we need to first identify steps that meet the necessary conditions for hallucination generation. To achieve this, we employ an external large language model and develop a set of tailored rules within the prompts. These rules guide the language model in determining whether a reasoning step in the context provides the elements required for a specific type of hallucination.
For example, when evaluating whether a step can introduce factual inconsistency, the language model checks if the reasoning step references objects (e.g., quantities, features) or named entities. This enables us to manipulate the information, allowing for the seamless integration of contextual inconsistencies in later steps. The complete set of rules for identifying hallucination injection position across the six hallucination types is detailed in Appendix~\ref{sec: judging prompt}.



\paragraph{Step 2: Hallucinate ground truth reasoning steps}
After confirming the appropriate position for injecting the hallucination, we present a mathematical problem and the correct reasoning history to an external large language model, instructing it to generate the next reasoning step with the target hallucination.
To control the distribution of hallucinations in the generated dataset and improve the success rate of incorporating our hallucination taxonomy, we prompt the language model to insert each type of hallucination separately.
We begin by inputting specific instructions for each hallucination type into the system prompt, guiding the language model to modify the reasoning process and introduce the desired hallucination. Detailed instructions for each hallucination type are provided in Appendix \ref{Synthesis prompt}.
Next, we employ an in-context learning strategy by providing two demonstrations for each query. Each demonstration includes an example of an injected hallucination, along with an explanation how it is introduced. These demonstrations can be found in Appendix \ref{Hallucination Demonstration}.
To reduce the financial cost, we delegate the task of hallucinating reasoning steps to the Llama-3-70B model~\citep{dubey2024llama}. We experimentally found that our method enables the language model to generate hallucinatory reasoning steps efficiently.
More details are in Appendix \ref{sec:other-generators}.


\subsubsection{Model Training}


After generating six types of hallucination datasets with our automated data annotation method, we train our FG-PRM, denoted as \(R_{\Phi}\), which comprises six PRMs, $R_{\phi_1} \dots R_{\phi_6}$, each corresponding to a specific type of hallucination in our taxonomy.

Formally, given an input question \(x\) and the corresponding solution \(y\) composed of \(L\) reasoning steps \(\{y_1, y_2, \dots, y_L\}\), we separately train task-specific PRMs $R_{\phi_t}$ to detect whether each reasoning step in $y$ contains the hallucination type $t$. The model input has the format of ``$\texttt{question: } q, \ \texttt{reasoning steps: } y_1\ \texttt{[sep] }y_2\\ \texttt{[sep]} \dots y_L\ \texttt{[sep]}$”, where each \texttt{[sep]} token represents the classification output at each reasoning step to indicate whether the previous step $y_i$ contains the hallucination type $t$.
We define $R_{\phi_t}(x, y_i)  = P(\texttt{[sep]} = 1)$ to represent the probability that the step $y_i$ contains the hallucination type $t$.
To train each PRM $R_{\phi_t}$, we utilize a step-level classification loss as in Eq.\ref{equa： PRM loss} to each \texttt{[sep]} token before step $y_i$. Overall, our FG-PRM \(R_{\Phi}\) generates an aggregate reward for the solution \(y\) of the input question \(x\):

\vspace{-5mm}
\begin{align}
    R_{\Phi}(x, y) = 
    \sum_{t=1}^{6} \sum_{i=1}^{L} \log \Big( R_{\phi_t} (x, y_i)\Big) 
\end{align}


In the verification task(see Appendix \ref{verification}), the log-sum of these probabilities is used to aggregate the rewards, resulting in the final reward assigned by FG-PRM for a solution. Importantly, if a step is correct, its probability of correctness is close to 1, contributing minimally to the aggregated reward. This ensures that the length of the answer does not influence the final reward score, maintaining fairness regardless of the sequence length.




\section{Experiments}
\subsection{Settings}

\paragraph{Task description}
We test our FG-PRM on two tasks: fine-grained hallucination detection and mitigation in language models. The detection task aims to identify specific types of hallucinations at each reasoning step, using precision, recall, and F1 scores to evaluate performance. The mitigation task involves ranking multiple candidate solutions for a problem, with a reward model assigning scores based on correctness to select the best solution. Detailed descriptions are presented in Appendix~\ref{task formulation}.

\paragraph{Datasets}
We conduct our experiments on two widely used mathematical benchmarks, GSM8K \citep{cobbe2021training} and MATH \citep{hendrycks2021measuring}.
GSM8K consists of grade school math problems designed to benchmark the reasoning abilities of language models.
To construct the hallucinatory reasoning steps, we employ a meta-dataset and software library \citep{ott2023thoughtsource}, which collects the golden chain-of-thought solutions for problems in the GSM8K.
MATH, on the other hand, is a large-scale dataset designed for probing and improving model reasoning, which includes human-written step-by-step solutions~\citep{lightman2023let}.

Following \citep{lightman2023let, uesato2022solving, wang2023math}, we randomly sample 700 instances from the training set and 100 instances from the test set for both GSM8K and MATH.
We refer to these datasets as ``Coarse-grained Hallucinations'' (CG-H), which include human-annotated reasoning chains with each step labeled for correctness.
Using the two sampled datasets, we augment each to 12,000 instances through our automatic hallucination generation method described in Section \ref{sec:auto-hallu-gen}, covering all types of hallucinations mentioned in Section \ref{sec:taxonomy} with a balanced hallucination distribution. These augmented datasets are termed ``Fine-grained Hallucinations'' (FG-H).
Furthermore, we sample 12,000 instances from Math-Shepherd \citep{wang2023math}, a dataset consisting of automatically constructed process-wise supervision data using the Monte Carlo tree search method, with each step labeled for correctness.

For the hallucination detection task, we construct both synthetic and human-annotated datasets based on questions from the MATH dataset. The CG-H (MATH) and FG-H (MATH) datasets are utilized to train PRM and FG-PRM, respectively, and to evaluate their performance. For the hallucination mitigation task, we employed the complete CG-H and FG-H datasets, along with Math-Shepherd, to train various reward models. All experiments maintain a training-to-validation split ratio of 95:5.

%

%

\begin{table*}[t!]
\centering
\footnotesize

\addtolength{\tabcolsep}{-1.8pt}  
\begin{tabular}{l|p{0.037\textwidth}p{0.037\textwidth}p{0.037\textwidth}p{0.037\textwidth}p{0.037\textwidth}p{0.037\textwidth}c|p{0.037\textwidth}p{0.037\textwidth}p{0.037\textwidth}p{0.037\textwidth}p{0.037\textwidth}p{0.037\textwidth}c}\toprule
& \multicolumn{7}{c|}{Synthetic Reasoning Chain} & \multicolumn{7}{c}{Human-annotated Reasoning Chain} \\
\midrule
Detector &
\CI &  \LI &  \II & \LE & \FI & \FA & Average& 
\CI &  \LI &  \II & \LE & \FI & \FA &Average\\\midrule
ChatGPT & 0.415 & 0.522 & 0.453 & 0.360 & \bf 0.428 & 0.900 & \bf 0.513 & 0.442 & 0.552 & 0.510 &\bf 0.377 & \bf 0.487 & \bf 0.840 & \bf 0.531  \\
Claude & 0.448 & 0.388 & 0.493 & 0.275 & 0.373 &\bf 0.963 & 0.490 & 0.434 & 0.460 & 0.478 & 0.359 & 0.428 & 0.758 & 0.503 \\ \midrule
PRM & 0.399 & 0.455 & 0.467 &\bf 0.402 & 0.358 & 0.565 & 0.441 & 0.394 & 0.493 & 0.484 & 0.357 & 0.403 & 0.435 & 0.428 \\
FG-PRM &\bf  0.488 & \bf 0.549 &\bf 0.529 & 0.398 & 0.422 & 0.608 & 0.499 &\bf 0.526 & \bf 0.575 &\bf 0.513 &\bf 0.377 & 0.426 & 0.484 & 0.484 \\
\bottomrule
\end{tabular}
\vspace{-2mm}
\caption{
Performance of fine-grained hallucination detection across all hallucination types on synthetic data and human-annotated data. All numbers are F1 scores.
}
\label{tab:results_automatic_detection}
\vspace{-5mm}
\end{table*}

\paragraph{Models}
In the fine-grained hallucination detection task, we evaluate the performance of prompt-based and model-based detection.
For prompt-based detection, we apply ChatGPT (GPT-3.5-turbo-0125)~\citep{ouyang2022training} and Claude (Claude-3-haiku)\footnote{\url{https://claude.ai/}} with carefully designed prompts as baseline methods.
For the model-based detection, we compare our FG-PRM with the traditional coarse-grained PRM~\citep{lightman2023let}.
In the fine-grained hallucination mitigation task, we apply various verifiers to evaluate the correctness of solutions generated by language models.
We employ Llama-3-70B~\citep{dubey2024llama} as our solution generator, from which we sample 64 candidate solutions for each test problem. We apply the LongFormer-base-4096~\citep{beltagy2020longformer} and Llama-3-8B~\citep{dubey2024llama} as our base models due to their strong performance in handling long-context reasoning.
We keep their main structure unchanged. Specifically, we replace the output layer with an MLP to predict binary hallucination labels for reward models.
%
Verifiers include self-consistency (SC), ORM, PRM, CG-PRM, FG-ORM, and FG-PRM.
The SC verifier serves as a baseline without specific model training. It aggregates multiple reasoning paths and selects the most frequent solution as the final answer.
ORMs and PRMs are trained on the CG-H dataset.
For CG-PRM, we train a single PRM on the coarsely labeled FG-H datasets, using binary labels like CG-H instead of fine-grained types.
For FG-ORM and FG-PRM, we train individual fine-grained ORMs and PRMs for each of six hallucination types on FG-H dataset, respectively.
All experiments are performed on four NVIDIA A100 80G GPUs.

\subsection{Hallucination Detection Results}

To evaluate the efficacy of our method in detecting fine-grained hallucinations, we conduct two experiments on synthetic and human-annotated data.

\paragraph{Synthetic Data} 

We utilize the automated annotation labels from our synthetic dataset, FG-H (MATH), as the golden standard for evaluating various detectors across six types of hallucination. As shown in Table \ref{tab:results_automatic_detection}, FG-PRM outperforms prompt-based detectors in detecting \CI, \LI, \II, and \LE, demonstrating FG-PRM has effectively learned the patterns of these hallucinations and can detect them accurately. 
Notably, FG-PRM outperforms PRM in detecting all types of hallucination, demonstrating the advantages of the fine-grained detection manner.
However, prompt-based detectors outperform FG-PRM on \FI\ and \FA, primarily due to their larger model sizes and greater access to fact-based knowledge. This reflects the inherent advantage of large language models in fact-based verification.
Moreover, detailed precision and recall results are presented in Tables \ref{tab: presition} and \ref{tab: recall} in Appendix \ref{sec: FG result}.
Besides the six individual verifiers in FG-PRM, we also conduct experiments on a single multi-class verifier, which performs worse than FG-PRM. Additional results are shown in Appendix \ref{sec: compact}.

\paragraph{Human-annotated Data}
\label{sec:human-annotation}


We also validate the effectiveness of our method on real-world data using human-annotated data. 
Specifically, for each hallucination type, we first utilize ChatGPT to generate step-by-step solutions for 50 problems from the MATH dataset. Five NLP experts then manually annotate these solutions according to our hallucination taxonomy, ensuring each selected solution includes at least one step exhibiting the target hallucination type. This process resulted in a human-labeled dataset covering six hallucination types, each represented by 50 annotated responses corresponding to 50 problems, featuring step-level hallucination labels.
The annotations achieved a Cohen’s Kappa score of 0.79, indicating substantial agreement among the annotators.


In Table~\ref{tab:results_automatic_detection}, results on the human-annotated data closely align with the trends observed on the synthetic data. FG-PRM demonstrates superior performance in detecting \CI\ and \LI\ hallucinations, consistently outperforming all other models in these categories. However, FG-PRM's performance is slightly below that of strong, non-public LLMs (e.g., ChatGPT and Claude) in detecting \FI\ and \FA\ hallucinations. This discrepancy is largely attributable to FG-PRM's smaller parameter size and limited access to world knowledge. Despite these challenges, FG-PRM performs competitively overall, particularly in reasoning-related hallucinations. Further analysis on reasoning chain evaluation for various verifiers is presented in Appendix \ref{sec: evaluation}.

\begin{table}[t]
\centering
\resizebox{\columnwidth}{!}{%
\begin{tabular}{llcc}
\toprule
Base Model                  & Verifier / Reward Model   & GSM8K & MATH\\
\midrule
-                           & Self-Consistency          & 0.88 & 0.48 \\
\midrule
\multirow{6}{*}{LongFormer} & ORM                       & $0.88^\dagger$ & 0.51 \\
                            & PRM                       & 0.89 & 0.53 \\
                            & Math-Shepherd (ORM)       & 0.90 & 0.52 \\
                            & Math-Shepherd (PRM)       & 0.91 & 0.54 \\
                            \cmidrule{2-4}
                            & CG-PRM (Ours)             &  0.89 & 0.54 \\
                            & FG-ORM (Ours)             &  0.89 & 0.53 \\
                            & FG-PRM (Ours)             &  \textbf{0.94} & \textbf{0.57} \\
\midrule
\multirow{6}{*}{Llama-3-8B}  & ORM                       & $0.87^\dagger$ & 0.52 \\
                            & PRM                       & 0.90 & 0.53 \\
                            & Math-Shepherd (ORM)       & 0.89 & 0.51 \\
                            & Math-Shepherd (PRM)       & 0.91 & 0.53 \\
                            \cmidrule{2-4}
                            & CG-PRM (Ours)             &  0.90 & 0.54 \\
                            & FG-ORM (Ours)             &  0.89 & 0.53 \\
                            & FG-PRM (Ours)             &  \textbf{0.93} & \textbf{0.58} \\
\bottomrule
\end{tabular}
}
\vspace{-2mm}
\caption{
Performance of different verifiers on GSM8K and MATH benchmarks.
The evaluation is based on 64 candidate solutions for each problem generated by Llama3-70B model with greedy decoding.
We calculate the mean of 3 groups of sampling results.
Statistical significance test indicates that most improvements compared to ``Self-Consistency'' are significant ($p< 0.05$).
Data marked with $^\dagger$ indicate the significant test with $p< 0.05$ is not passed.
More results are in Appendix \ref{sec:data_scaling_analysis}.
}
\label{tab: verification task}
\vspace{-7mm}
\end{table}

\begin{figure*}[ht]
    \vspace{-7mm}
    \centering
    \begin{minipage}[b]{0.45\linewidth}
        \centering
        \includegraphics[width=\linewidth]{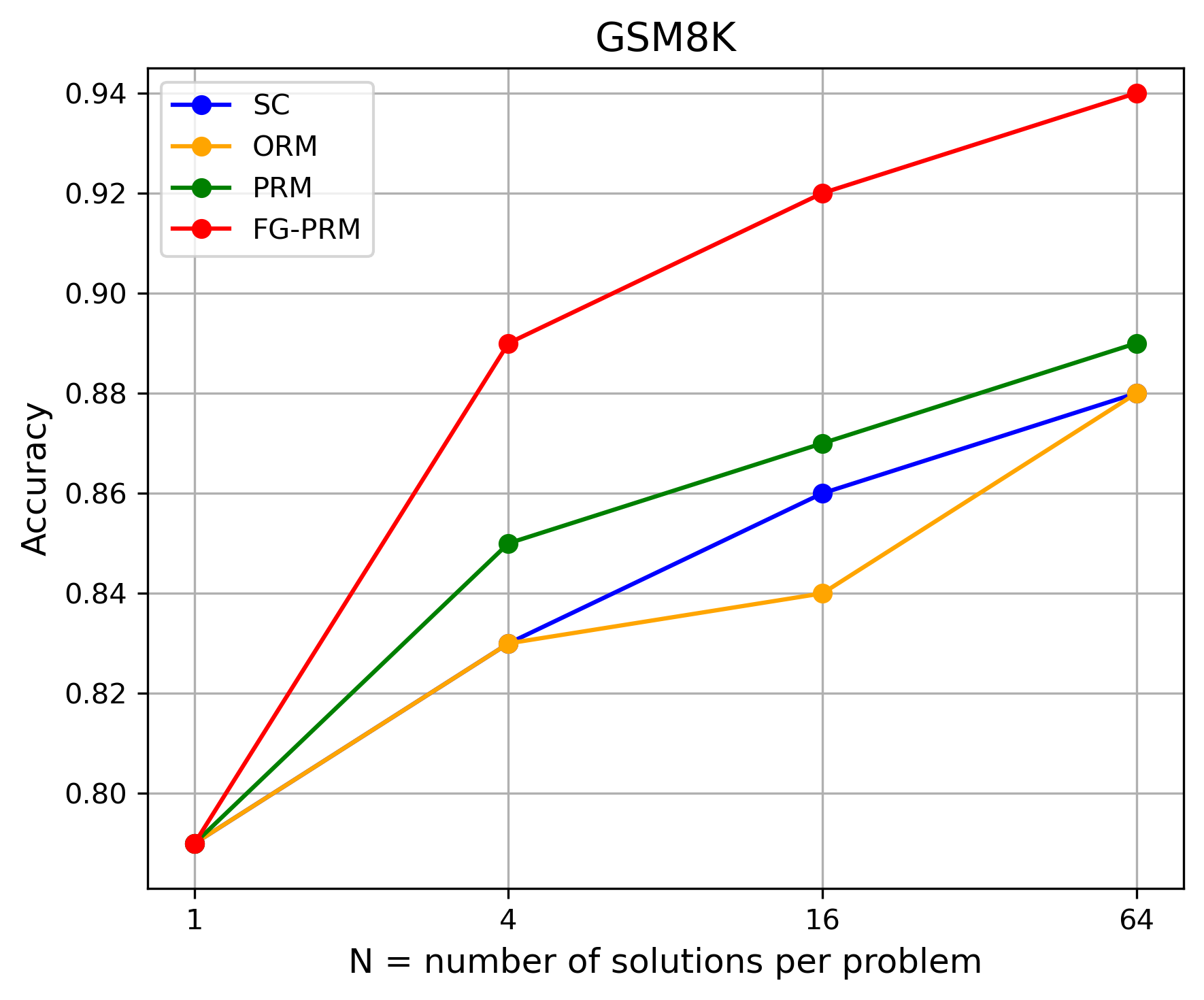}
        \label{fig:gsm8k}
    \end{minipage}
    \hspace{0.05\linewidth}
    \begin{minipage}[b]{0.45\linewidth}
        \centering
        \includegraphics[width=\linewidth]{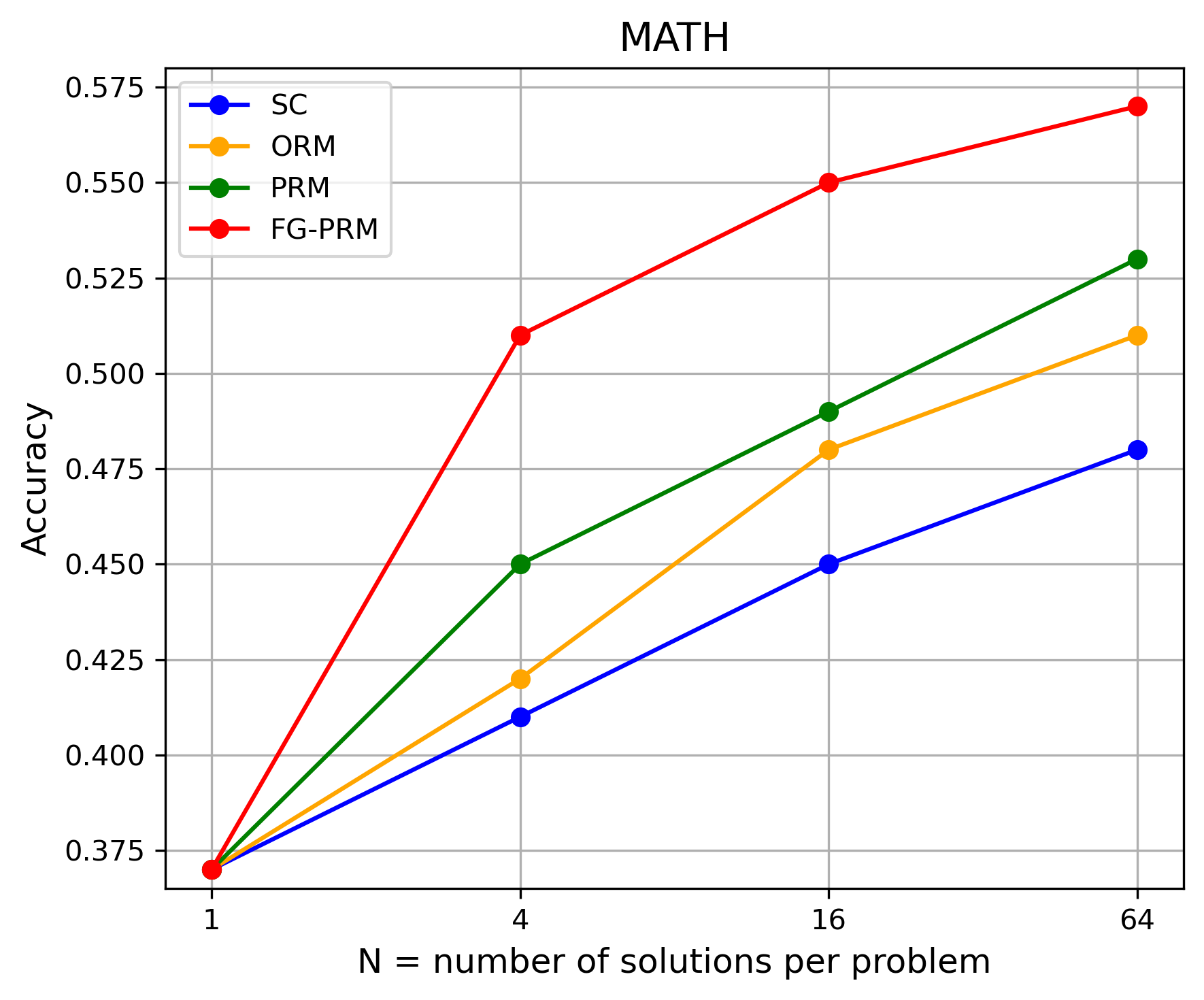}
        \label{fig:math}
    \end{minipage}
    \vspace{-8mm}
    \caption{Performance of LLama-3-70B across varying candidates on GSM8K and MATH using different verifiers.}
    \label{fig:combined}
    \vspace{-5mm}
\end{figure*}

\subsection{Hallucination Mitigation Results}
Table \ref{tab: verification task} presents a performance comparison of various verifiers on GSM8K and MATH.
FG-PRM trained on our augmented dataset, FG-H, significantly outperforms all baselines across both base models.
Specifically, after fine-tuning with FG-H, Longformer and Llama3-8B achieve 94\% and 58\% accuracy on GSM8K and MATH, respectively, surpassing PRMs trained on Math-Shepherd.
The results show that base models mitigated by PRMs consistently outperform those mitigated by ORMs, consistent with findings from \cite{uesato2022solving,lightman2023let,wang2023math}.
On GSM8K, most baseline verifiers perform close to the self-consistency level due to the simplicity of the dataset, where many questions involve only basic arithmetic operations. 
However, the differences between verifiers become more evident in the more complex MATH dataset, where questions and reasoning steps often require LaTeX math expressions.
Comparing PRM and CG-PRM, increasing the training size alone does not yield significant improvements.
Moreover, the enhancements in FG-ORM over ORM and FG-PRM over CG-PRM demonstrate the effectiveness of our fine-grained approach.
Notably, FG-ORM and CG-PRM, trained on the same data size as FG-PRM, are inferior to FG-PRM. 
These results indicate that the balanced fine-grained step-level supervision employed by FG-PRM offers a more robust and effective approach to hallucination mitigation, particularly in handling complex problem-solving tasks.

\vspace{-1mm}

\section{Analysis}

\paragraph{Hallucination Mitigation Performance with Varying Candidate Solutions}

Figure~\ref{fig:combined} illustrates the performance of four verifiers with the number of candidate solutions ranging from 1 to 64 across two benchmarks.
This demonstrates that FG-PRM consistently outperforms all other verifiers.
With predicted insights, the performance gap between FG-PRM and other baseline verifiers will increase with the growth of $N$.



\paragraph{Out-of-Distribution Dataset Evaluation}

We further conduct out-of-distribution (OOD) evaluation experiments to assess the robustness and transferability of our approach.
In these experiments, we train the PRM verifier on CG-H (GSM8K) and FG-PRM verifier on FG-H (GSM8K). For comparison, we also train them on CG-H (MATH) and FG-H (MATH). Finally, we test all verifiers on the MATH dataset with 64 candidate solutions for each questions.
Notably, the GSM8K dataset contains simple questions, predominantly solvable through basic arithmetic operations, in contrast to the more complex MATH dataset.


\begin{figure}[h]
    \vspace{-4mm}
    \centering
    \footnotesize
    \includegraphics[width=\columnwidth]{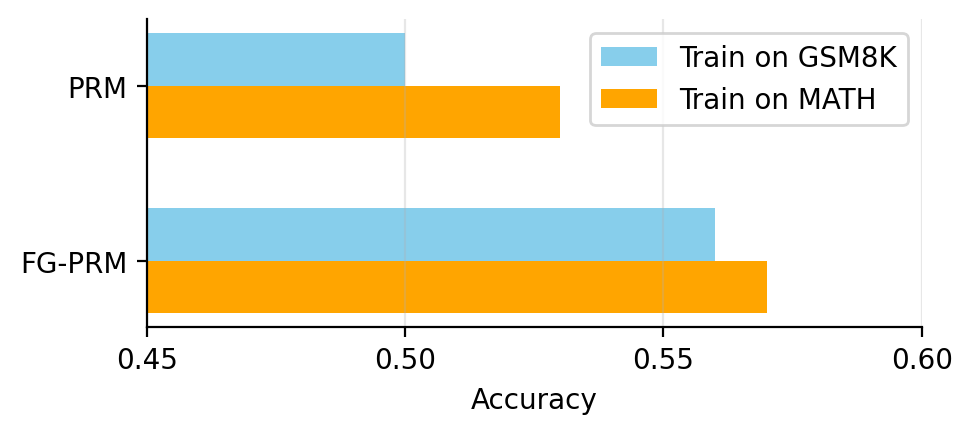}
    \vspace{-8mm}
    \caption{OOD performance on the MATH benchmark.}
    \label{fig:out-of-distribution}
    \vspace{-3mm}
\end{figure}

\begin{table*}[ht]
\setlength\tabcolsep{4pt}
\small
\vspace{-5mm}
\centering
    \begin{tabular} {p{1.5cm}|p{13.8cm}}
        \toprule
         \textbf{Question} & \textit{Each edge length of a rectangular solid is a prime number. If the volume of the rectangular solid is 385 cubic units, what is the total surface area, in square units, of the rectangular solid?}
         \\
         \midrule
      \textbf{Answer} & {334} \\ \midrule
  \textbf{Reasoning Steps}       & Step 1:We know that the volume of this rectangular solid is the product of its three side lengths.   \textcolor{teal}{FG-PRM: 0.95}, \textcolor{olive}{CG-PRM: 0.83}. \\
         & Step 2: Let's call those side lengths $a$, $b$, and $c$. The volume is $abc$. \textcolor{teal}{FG-PRM: 0.83}, \textcolor{olive}{CG-PRM: 0.80}.\\
         & ...... \\
         & Step 6: Let's plug in the possible values. If $a=5$, $b=7$, and $c=11$, then $2ab+2bc+2ac=2(5)(7)+2(7)(11)+2(5)(11)=74+154+110=338$. \textcolor{teal}{FG-PRM: 0.87}, \textcolor{olive}{CG-PRM: 0.72}. \\
         & Step 7: Which simplifies to $14\cdot{11}+22\cdot{11}+10\cdot11$. \textcolor{teal}{FG-PRM: \li\ 0.07} \textcolor{olive}{CG-PRM: 0.80}. \\
          & Step 8: So the total surface area would be $334$. \textcolor{teal}{FG-PRM : 0.88} \textcolor{olive}{CG-PRM: 0.66}. \\
         \bottomrule
    \end{tabular}
    \vspace{-2mm}
    \caption{
    A case study from the MATH dataset.
    A high \textcolor{olive}{CG-PRM} score indicates that the step is positive.
    A low \textcolor{teal}{FG-PRM-TYPE} score indicates the step has a high probability of having the TYPE of hallucination.
    }
    \label{tab:case}
    \vspace{-5mm}
\end{table*}

As detailed in Figure \ref{fig:out-of-distribution}, the performance gap for FG-PRM ({+0.01}) is slighter than PRM ({+0.03}).
Moreover, the verifiers trained on the FG-H (GSM8K) demonstrate performance closely comparable to those trained on the FG-H (MATH) dataset.
This indicates that the FG-PRM verifier trained on FG-H can learn to recognize patterns of hallucinations and generalize this knowledge to tackle more challenging scenarios effectively.

\paragraph{Qualitative Result of Fine-Grained Hallucination Evaluation}

To illustrate the effect of our FG-PRMs, we demonstrate a case study in Table~\ref{tab:case}, comparing the hallucination detection results of FG-PRM and CG-PRM.
Our FG-PRM demonstrates remarkable discrimination by precisely detect fine-grained types of hallucination in reasoning steps.
%
Notably, Step 7 has a high \li\ score since the step incorrectly refers to the formula in previous steps.
It refers to $a=5$ as $a=11$, which causes \li. FG-PRM recognizes it by giving a low score, while CG-PRM does not recognize this detail.

\section{Related Work}


\paragraph{Hallucinations in LLM} Numerous studies have explored the taxonomy of hallucinations in language models. \citet{mishra2024fine} identify six fine-grained types of factual hallucinations in information search scenarios. \citet{zhang2023siren} classify hallucinations based on conflict types: input-conflicting, context-conflicting, and fact-conflicting. \citet{huang2023survey} categorize hallucinations into factuality and faithfulness types and divide them further. \citet{ji2023survey} discussed intrinsic and extrinsic hallucinations, focusing on whether outputs contradict input content. 
Closest to our work, \citet{golovneva2022roscoe} propose ROSCOE, measuring semantic alignment, similarity, and language coherence in reasoning chains. Unlike ROSCOE that includes many grammatical errors like grammar, redundancy, and repetition, our taxonomy provides detailed distinctions between error types for diagnosing complex reasoning errors and improving model outputs.


\paragraph{Evaluation of Reasoning Chains}
Depending on whether golden references are required, methods to evaluate reasoning chains can be roughly divided into reference-dependent and reference-free ones. 
For reference-dependent, the reasoning chains can be evaluated with LLMs~\citep{ren2023investigating, adlakha2023evaluating}), or by measuring the discrepancy between the vanilla response and reference \citep{huo2023retrieving, pezeshkpour2023measuring}. 
For reference-free metrics, some methods rely on aggregating the individual token probabilities assigned by the LLM during generation so that they can reflect reasoning chain uncertainty~\citep{manakul2023selfcheckgpt, huang2023look}. 
In addition to that, many model-based methods have emerged to evaluate reasoning chains~\citep{he2024socreval, hao2024llm}. 
In this work, we focus on model-based reference-free reasoning chain evaluation from the perspective of hallucination detection.

\paragraph{Improving reasoning abilities of LLMs} For LLMs that have completed training, prompting techniques are an effective approach to improve the performance of LLMs on reasoning tasks without modifying the model parameters~\citep{wei2022chain, fu2022complexity, yao2024tree}. 
Besides, instead of directly improving the reasoning performance of LLMs, verifiers, typically the Outcome Reward Model (ORM) and Process Reward Model (PRM), can raise the success rate in solving reasoning tasks by selecting the best answer from multiple decoded candidates.
PRM provides a more detailed evaluation by scoring each step. However, training a PRM requires access to expensive human-annotated datasets. Methods such as Math-Shepherd \citep{wang2023math} and MiPS \citep{wang2024multi} have explored Monte Carlo estimation to automate the data collection process without human involvement, and OmegaPRM \citep{luo2024improve} proposed a divide-and-conquer style Monte Carlo tree search algorithm for automated process supervision data generation. 

\section{Conclusion }

In conclusion, we introduce FG-PRM, a nuanced approach for detection and mitigation of hallucinations in language model reasoning. We proposed a taxonomy to categorize hallucinations into six types.  By leveraging a novel automated data generation method, we significantly reduce the dependency on costly human annotations while enriching the dataset with diverse hallucinatory instances. 
Our empirical results show that FG-PRM, when trained on our synthetic data, significantly enhances the accuracy of hallucination detection, providing an effective approach for improving the LLM reasoning accuracy and faithfulness.

\paragraph{Limitation} Our automated data synthesis approach depends on ground-truth reasoning step datasets, which may pose a constraint for large-scale scaling. Additionally, this approach relies on running LLM APIs or performing inference using GPUs to generate hallucinations. While effective, this method is not the most cost-efficient. Future work could focus on developing more efficient solutions to reduce costs and improve scalability. In future work, we plan to extend FG-PRM from mathematical reasoning task to other domains (e.g., scientific QA, commonsense reasoning).

\section*{Acknowledge}
We thank the anonymous reviewers for valuable and insightful feedback.
This research is supported in part by the National Science Foundation CAREER Grant IIS-2340435 and an Amazon Research Award.  Any opinions, findings, and conclusions or recommendations expressed herein are those of the authors and do not necessarily represent the views, either expressed or implied, of the U.S. Government.

\bibliography{custom}

\appendix
\label{sec:appendix}
\clearpage
\appendix
\onecolumn
\section{Examples of Fine-Grained Hallucination}

\begin{table*}[ht]
  \vspace{-3mm}
  \centering
  \footnotesize
  \caption{Examples of various types of LLM hallucinations in our taxonomy. Content marked in \textcolor{red}{Red} represents the hallucinatory output, while content marked in \textcolor{blue}{Blue} indicates user instruction or provided context that contradicts the LLM hallucination.}
  \renewcommand{\arraystretch}{1.3}
  \vspace{1mm}
  \resizebox{\textwidth}{!}{
    \begin{tabular}{p{2cm}p{2cm}p{3cm}p{3cm}p{4cm}}
    \toprule
    \textbf{Hallucination Type} & \textbf{Sub-Type} & \textbf{User Input} & \textbf{Model Output} & \textbf{Explanation}\\
    \midrule
    \multirow{10}[2]{*}[-6em]{\makecell[{{p{3cm}}}]{Extrinsic \\Hallucinations}} & \makecell[l]{Factual \\ Inconsistency} & What is the capital of Australia? & The capital of Australia is \textcolor{red}{Sydney}. & The model incorrectly states that Sydney is the capital of Australia, while the correct capital of Australia is Canberra. This error demonstrates factual inconsistency as the model’s response contradicts the real-world fact. \\
    \cmidrule{2-5}
          & \makecell[l]{Fabrication} &What is the population of Atlantis? & \textcolor{red}{The population of Atlantis is approximately 1.2 million people.} & The model provided a specific population figure for Atlantis, a fictional place with no basis in reality. Since Atlantis does not exist, there is no verifiable population data.\\
    \cmidrule{2-5}
          & \makecell[l]{Calculation \\ Error} &  If a rectangle has a length of 4 units and a width of 3 units, what is the perimeter of the rectangle? & The perimeter formula of a rectangle is: Perimeter = 2 × (length + width). The length is 4 units and the width is 3 units. \textcolor{red}{So, the perimeter would be 2 × (4 + 3) = 16.} So the answer is 16 units. & In the red reasoning step, the model incorrectly calculates 2 × (4 + 3) = 16, which should be 14. The error should be verified by external tools like calculators, which matches the definition of Calculation Error hallucination.\\
    \midrule
    \multirow{2}[3]{*}[-12em]{\makecell[{{p{3cm}}}]{Intrinsic \\ Hallucination}} & \makecell[l]{Instruction \\ Inconsistency}  & Please calculate the quotient of 25 divided by 100 and \textcolor{blue}{represent the result as a fraction}. & The quotient of 25 divided by 100 is \textcolor{red}{0.25}. & Although the model correctly calculates the result 0.25, it fails to follow the ``as a fraction” instruction in the user input. \\ 
    \cmidrule{2-5}
          & \makecell[l]{Context \\ Inconsistency} & Given that I have 5 apples and \textcolor{blue}{I give 2 to my friend}, how many apples do I have left? & Considering the information provided, if you initially had 5 apples and \textcolor{red}{you gave 3 to your friend}, you would have 2 apples left. & 
          In the red reasoning step, the model incorrectly refers to the number of apples in the context. The user stated that they gave 2 apples to their friend, but the model incorrectly refers to it as 3.\\
    \cmidrule{2-5}
          & \makecell[l]{Logical \\ Inconsistency} &  How many hours are there in a week if each day has 24 hours? & There are \textcolor{red}{7 days in a week}. Since each day has 24 hours, \textcolor{red}{a week would have 6 days} times 24 hours, which equals 144 hours. &  The model correctly states that there are 7 days in a week but then incorrectly calculates the number of hours by multiplying 6 days by 24 hours to get 144 hours. This shows a contradiction between the days in a week referred and a failure to maintain internal logical consistency in the reasoning process. \\
    \bottomrule
    \end{tabular}
    } 
  \label{tab:hallucination_example}
  \vspace{-10mm}
\end{table*}%

\onecolumn

\section{Task Formulation}
\label{task formulation}

In this section, we elaborate on the two primary tasks of fine-grained hallucination detection and mitigation. 

\subsection{Task 1: Fine-Grained Hallucination Detection}


This task aims to detect hallucinations in language model reasoning output at a granular level, focusing on individual reasoning steps. Specifically, the detector is tasked with identifying fine-grained hallucinations in the output of a language model by assigning reward scores for each intermediate step in a reasoning chain. The objective is to classify hallucination types at the step level, determining whether a specific hallucination type is present.

Given a question $x$ and its solution \( y \) consisting of \( L \) reasoning steps, we assume the ground-truth annotations for hallucination types are available. These annotations, denoted as \( y_{i}^{*t} \in \{\text{TRUE}, \text{FALSE}\} \), provide a binary label for each hallucination type \( t \) at the \( i \)-th step, indicating whether the hallucination $t$ is present (TRUE) or absent (FALSE). The detector models predict \( y_{i}^{t} \), where \( y_{i}^{t} \) is the model’s predicted label for the \( i \)-th step and hallucination type \( t \). We evaluate the model’s performance using standard metrics for classification as in previous work~\citep{feng-etal-2023-factkb, mishra2024fine}: precision and recall. For each hallucination type \( t \), the precision measures the proportion of correct predictions out of all predictions where the model indicated the presence of a hallucination at a step, while recall measures the proportion of actual hallucination steps that the model correctly identified. These are computed as follows:
\begin{align}
\operatorname{Precision}^{t} & = \frac{\sum_{i \in L} \mathbb{I}\left[y_{i}^{t}=y_{i}^{*t}\right]}{\sum_{i \in L} \mathbb{I}\left[y_{i}^{t}=\text{TRUE}\right]} \\
\operatorname{Recall}^{t} & = \frac{\sum_{i \in L} \mathbb{I}\left[y_{i}^{t}=y_{i}^{*t}\right]}{\sum_{i \in L} \mathbb{I}\left[y_{i}^{*t}=\text{TRUE}\right]}
\end{align}

Here, \( \mathbb{I} \left[ \cdot \right] \) is an indicator function that returns 1 if the condition is true and 0 otherwise. Precision indicates the proportion of correctly predicted hallucinations for type \( t \), while recall indicates how many of the true hallucinations were detected by the model.

To assess the overall performance across all hallucination types, we calculate the F1 score, which is the harmonic mean of precision and recall. The F1 score is computed for each hallucination type and then averaged across all types \( \mathcal{E} \):
\begin{align}
    \text{F1 Score} = \frac{1}{|\mathcal{E}|} \sum_{t \in \mathcal{E}} \frac{2 \times \operatorname{Precision}^{t} \times \operatorname{Recall}^{t}}{\operatorname{Precision}^{t} + \operatorname{Recall}^{t}}
\end{align}

Thus, fine-grained hallucination detection can be framed as a set of binary classification tasks, where the system predicts whether each reasoning step \( s_i \) contains a specific hallucination type. By evaluating precision, recall, and F1 score across different hallucination types, we gain a comprehensive understanding of the model's ability to detect and categorize hallucinations within complex reasoning processes.



\subsection{Task 2: Fine-Grained Hallucination Mitigation}
\label{verification}
The verification task~\citep{lightman2023let} assesses a model’s ability to evaluate and rank multiple candidate solutions for a given problem. In this task, a generator produces \(N\) possible solutions \(\{y^1, y^2, \dots, y^N\}\) for a problem \(x\), which are then evaluated by a reward model (Section \ref{sec:rm_prelim}). The reward model assigns a score to each candidate solution based on its correctness, with the goal of selecting the best solution among the candidates.

This task follows the best-of-N selection method, where the solution with the highest score is chosen as the final answer. A well-performing reward model improves the likelihood of selecting the correct solution, thereby enhancing the overall problem-solving accuracy. By providing meaningful feedback on each candidate solution, the verification task helps ensure that the reasoning process is grounded in correctness and consistency. 

\section{Detailed Fine-grained Hallucination Detection Results}
\label{sec: FG result}
The precision and recall of the fine-grained detection results for the Llama3-70B generation are reported in Table~\ref{tab: presition} and \ref{tab: recall}, respectively.

\begin{table*}[htbp]
\centering
\begin{tabular}{c|cccccc|c}
\toprule
 & \multicolumn{5}{c}{Hallucination Type} &\\
\midrule
Detector &  \CI &  \LI &  \II & \LE & \FI & \FA &  Average \\
\midrule
ChatGPT  & 0.403     & 0.488     & 0.450     & 0.424     & \bf 0.412 & {0.890}   & \bf 0.511    \\
Claude-3 & 0.417     & 0.368     & 0.490     & 0.248     & 0.357 & \bf 0.952 & 0.472     \\
PRM & 0.393 & 0.421 & 0.443 & 0.324 & 0.374 & 0.527 & 0.414 \\
FG-PRM   & \bf 0.428 & \bf 0.513 & \bf 0.528 & \bf 0.413 & 0.403 & 0.589     & 0.479 \\
\bottomrule
\end{tabular}
\caption{Precision for fine-grained hallucination detection across different categories.}
\label{tab: presition}
\end{table*}

\begin{table*}[htbp]
\centering
\begin{tabular}{c|cccccc|c}
\toprule
 & \multicolumn{5}{c}{Hallucination Type} &\\
\midrule
Detector &  \CI &  \LI &  \II & \LE & \FI & \FA &  Average \\
\hline
ChatGPT  & 0.440     & 0.600     & 0.460     & 0.541     & \bf0.477 & {0.920}   & \bf 0.573 \\
Claude-3 & 0.525     & 0.433     & 0.500     & 0.334     & 0.416    & \bf 0.990 & 0.533     \\
PRM & 0.415 & 0.498 & 0.493 & 0.541 & 0.352 & 0.615 & 0.486 \\
FG-PRM   & \bf 0.571 & \bf 0.597 & \bf 0.560 & \bf 0.546 & 0.462    & 0.635     & 0.562     \\
\bottomrule
\end{tabular}
\caption{Recall for fine-grained hallucination detection across different categories.}
\label{tab: recall}
\end{table*}

\section{Compact FG-PRM Verifier}
\label{sec: compact}

Besides six individual binary classifier verifiers in our FG-PRM, we train a single multi-class verifier on a Longformer model, denoted as FG-PRM (compact).
For this reward model, we replace the output layer with an MLP layer that produces seven category outputs, covering six types of hallucinations and a "no error" category.
As shown in Tables \ref{tab: compact} and \ref{tab: compact evaluation}, the performance of our separate FG-PRMs surpasses that of the compact experimental setting in both the hallucination detection and mitigation tasks.

Our findings indicate a tendency for the model to predict the "no error" label. This bias is primarily due to the imbalance in the training data.
Instances with a specific type of hallucination account for only one-sixth of the entire dataset.
Additionally, in the multi-step reasoning process, only a few steps display specific hallucinations.
As a result, models can achieve high accuracy by predominantly predicting "no error."

\begin{table*}[htbp]
\centering
\begin{tabular}{c|cccccc|c}
\toprule
 & \multicolumn{5}{c}{Hallucination Type} &\\
\midrule
Detector &  \CI &  \LI &  \II & \LE & \FI & \FA &  Average \\
\midrule
FG-PRM (Compact)  & 0.402                 & 0.493                 & 0.481                     & 0.378         & 0.371                    & 0.574       & 0.450   \\
FG-PRM            & 0.488                 & 0.549                 & 0.529                     & 0.398         & 0.422                    & 0.608       & 0.499  \\
\bottomrule
\end{tabular}
\caption{F1 score for fine-grained hallucination detection across different categories.}
\label{tab: compact}
\end{table*}

\begin{table}[htbp]
\centering
\vspace{1mm}
\resizebox{0.5\columnwidth}{!}{%
\small
\begin{tabular}{lcc}
\hline
Verifier / Reward Model & GSM8K         & MATH          \\ \hline
Self-Consistency        & 0.88          & 0.48          \\ \hline
FG-PRM (Compact)        & 0.90          & 0.54          \\
FG-PRM (Ours)           & \textbf{0.93} & \textbf{0.58} \\ \hline
\end{tabular}%
}
\caption{
Performance of FG-PRM and FG-PRM Compact verifiers on GSM8K and MATH benchmarks.
The evaluation is based on 64 candidate solutions generated by Llama3-70B model with greedy decoding for each problem.
Each result is the mean of results from 3 groups of sampling results.
}
\label{tab: compact evaluation}
\end{table}

\newpage
\section{Reasoning Step Hallucination Evaluation}
\label{sec: evaluation}

We utilize our model to evaluate hallucination issues in the generated outputs of large language models.
Each generation is assigned six scores corresponding to hallucination types.
The score under each hallucination type for a model is calculated based on the proportion of correct reasoning steps in generations. Specifically,
    $\text{score}=\frac{1}{N} \sum_{i=1}^N \frac{\text{\# of correct step}}{\text{\# of total step}}$
, where $N$ is the total number of generations in the test set. A model with high scores indicates fewer hallucination issues in its generation.

\begin{figure}[htbp]
    \centering
    \includegraphics[width=0.9\linewidth]{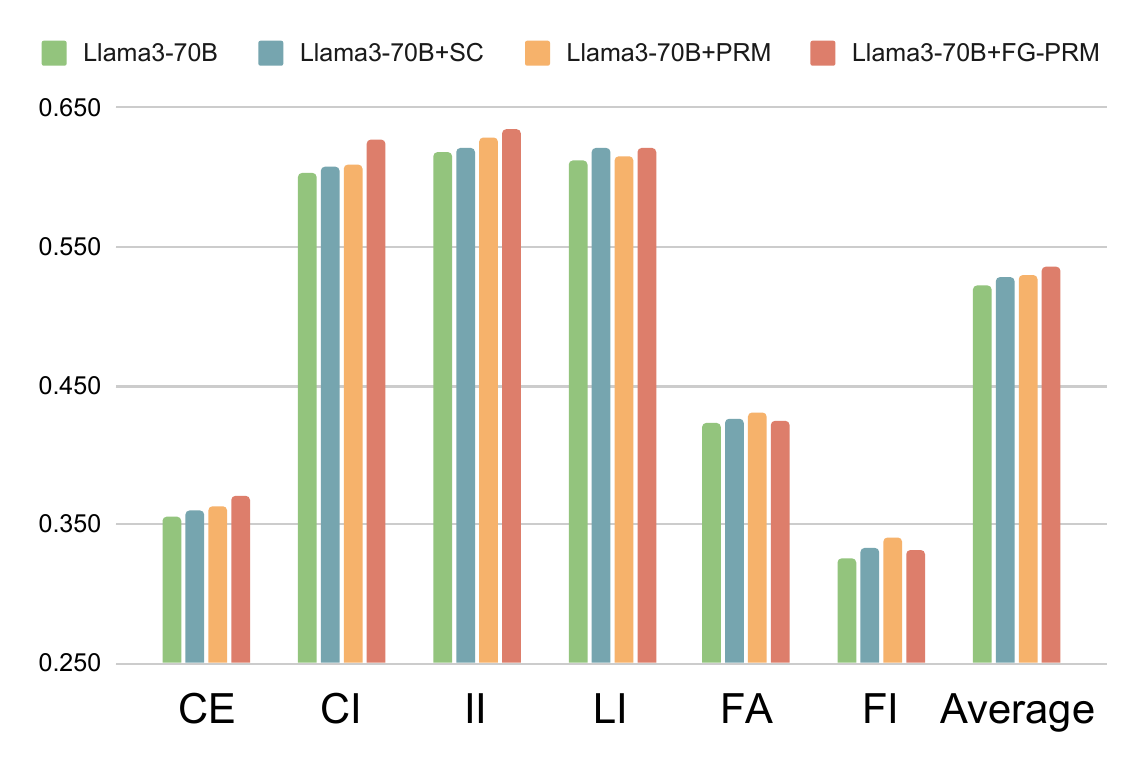}
    \caption{Hallucination evaluation performance on various models with verifiers.}
        \label{fig:benchmarking}
\end{figure}

Similar to the hallucination mitigation task, we apply our verifiers on Llama3-70B to help it select the best generation among 64 options.
The performance is shown in Figure \ref{fig:benchmarking}.
Llama3-70B, with help from verifiers, performs better than itself. The performance trend under each hallucination type aligns well with the results in Table \ref{tab: verification task} that FG-PRM performs the best among all verifiers.

\section{Generators}
\label{sec:other-generators}
In our experiments, we conducted a preliminary comparison of hallucination injection success rates using GPT-4o, Claude-3.5-Sonnet, and Llama3-70B on a balanced dataset with 50 examples per hallucination type. The success rates were comparable across models, aligning closely with the results presented in the table responding to question 4. While GPT-4o and Claude-3.5-Sonnet may yield slightly higher injection quality, the marginal gains are outweighed by their significantly higher API costs, estimated at over \$400 for generating 12K instances, which makes them impractical for large-scale annotation under our current budget constraints.

\section{More Analysis}
\label{sec:data_scaling_analysis}
\subsection{Data Scaling Analysis}
For the impact of the training data size, we conducted an empirical comparison between our FG-PRM trained on 12k instances and the PRM from Math-Shepherd trained on the full 400k dataset. Specifically, based on the existing results in Table \ref{tab: verification task}, we also evaluated the publicly released Math-Shepherd PRM model (Mistral-7B-based) using the same evaluation protocol described in their paper \cite{wang2023math}, with the minimum step score used as the reward. We repeated each evaluation 7 times and report the maximum scores in Table \ref{tab:data_scaling_analysis}.

These results show that:
(1) Despite using $\sim$30× fewer training examples, FG-PRM outperforms Math-Shepherd’s PRM 400K on both GSM8K and MATH benchmarks (+4\% on GSM8K with LongFormer, +2\% on MATH with Llama-3-8B).
(2) FG-PRM shows consistent performance gains across diverse base models. Notably, FG-PRM based on LongFormer (a smaller model) training on 12K data points still surpasses Math-Shepherd’s PRM training on 400K data points based on the more powerful Mistral-7B.
Moreover, it implicates that:
(1) FG-PRM’s fine-grained supervision (step-level error detection) enables it to learn more effectively from fewer examples.
(2) While Math-Shepherd relies on large-scale, coarse-grained supervision, our FG-H dataset offers high-quality, targeted hallucination patterns with balanced distributions.
These results underscore that FG-PRM’s performance stems primarily from its design and supervision strategy, rather than dataset scale alone. 

\begin{table}[ht]
\centering
\begin{tabular}{llcc}
\toprule
\textbf{Base Model} & \textbf{Verifier/ Reward Model} & \textbf{GSM8K} & \textbf{MATH}         \\
\midrule
LongFormer          & FG-PRM (Ours, 12K)            & \textbf{0.94} & 0.57          \\
                    & Math-Shepherd (12K)           & 0.91          & 0.54          \\
\midrule
Llama-3-8B          & FG-PRM (Ours, 12K)            & 0.93          & 0.58 \\
                    & Math-Shepherd (12K)           & 0.91          & 0.53          \\
\midrule
Mistral-7B          & Math-Shepherd (400K)          & 0.90          & 0.56 \\
\midrule
Qwen2.5-Math-7B     & FG-PRM (Ours, 12K)            & \textbf{0.94} & \textbf{0.60} \\
\bottomrule
\end{tabular}
\caption{Performance comparison between our FG-PRM and Math-Shepherd PRM on GSM8K and MATH benchmarks.}
\label{tab:data_scaling_analysis}
\vspace{-5mm}
\end{table}

\subsection{More Generator and Base Models Analysis}
We integrate Qwen2.5-Math-7B, a newer and stronger math-specific model, into our framework. In table \ref{tab:data_scaling_analysis}, Qwen2.5-Math-7B achieves comparable or superior results to LongFormer and Llama-3-8B on GSM8K and MATH benchmarks. This demonstrates FG-PRM’s adaptability to stronger mathematical architectures.

Moreover, we add the result of Qwen2.5-Math-7B as a solution generator, and compare it with existing results in Table \ref{tab:new_generator}.
With LongFormer-based FG-PRM, Qwen2.5-Math-7B as a generator improves MATH performance by +5\% compared to Llama-3-70B (0.57 $\rightarrow$ 0.62). With Llama-3-8B based FG-PRM, Qwen2.5-Math-7B as a generator improves MATH performance by +6\% compared to Llama-3-70B (0.58 $\rightarrow$ 0.64).
These results indicate that: (1) FG-PRM’s performance gains stem from its fine-grained supervision, and will benefit from stronger base models; (2) FG-PRM would likely benefit further from stronger math-specific generators like Qwen2.5-Math-7B.

\begin{table}[ht]
\centering
\resizebox{0.6\columnwidth}{!}{%
\begin{tabular}{lll}
\toprule
\textbf{Generator} & \textbf{Base model (FG-PRM)} & \textbf{MATH} \\
\midrule
Llama-3-70B        & LongFormer                   & 0.57          \\
Llama-3-70B        & Llama-3-8B                   & 0.58          \\
\midrule
Qwen2.5-Math-7B    & LongFormer                   & \textbf{0.62} \\
Qwen2.5-Math-7B    & Llama-3-8B                   & \textbf{0.64} \\
\bottomrule
\end{tabular}%
}
\caption{Performance comparison between two generators.}
\label{tab:new_generator}
\end{table}

\subsection{Ablation Study}
Based on the existing result in Table \ref{tab: verification task}, we systematically removed each hallucination type from FG-PRM and evaluated its impact on verification performance. The results are in table \ref{tab:ablation-study}. ``-{hallucination type}'' indicates that we only remove the PRM for that type of hallucination based on the FG-PRM when performing the verification task.

\begin{table}[ht]
\centering
\resizebox{0.7\columnwidth}{!}{%
\begin{tabular}{llll}
\toprule
\textbf{Base Model} & \textbf{Verifier /Reward Model} & \textbf{GSM8K}      & \textbf{MATH} \\
\midrule
Llama-3-8B          & FG-PRM                          & \textbf{0.93}       & \textbf{0.58} \\
                    & - Context Inconsistency         & 0.90 (-3\%)          & 0.55 (-3\%)    \\
                    & - Logical Inconsistency          & \textbf{0.90 (-3\%)} & 0.54 (-4\%)    \\
                    & - Instruction Inconsistency      & \textbf{0.92 (-1\%)} & 0.56 (-2\%)    \\
                    & - Factual Inconsistency          & 0.90 (-3\%)          & 0.55 (-3\%)    \\
                    & - Fabrication                    & 0.91 (-2\%)          & 0.56 (-2\%)    \\
                    & - Calculation                    & 0.89 (-4\%)          & 0.53 (-5\%)    \\
\bottomrule
\end{tabular}%
}
\caption{Ablation study. “- hallucination type” indicates that we only remove the PRM for that type of hallucination based on the FG-PRM when performing the verification task.}
\label{tab:ablation-study}
\end{table}

Based on the results, we can find that (1) Calculation Error has the largest impact (4\% drop on GSM8K and 5\% drop on MATH), aligning with its prevalence in math tasks. (2) Less Frequent Types (e.g., Instruction Inconsistency, Fabrication) still contribute to performance (1–3\% drops), suggesting their necessity for comprehensive error detection.

While mathematical reasoning is symbolic, it often intersects with real-world knowledge (e.g., units, constants, contextual facts). For instance:
(1) Factual Inconsistency: A model might incorrectly state that ``1 mile = 1.5 kilometers'' (instead of 1.609) or misattribute historical origins of mathematical theorems.
(2) Fabrication: An LLM could invent non-existent formulas (e.g., ``Euler’s Third Theorem'') to justify a step.
(3) Instruction Inconsistency: A model might ignore explicit problem constraints (e.g., ``calculating the area of a circle instead of the radius instructed in the prompt''). These errors, though less frequent than calculation errors, critically undermine reasoning validity. Our taxonomy aims to holistically capture failure modes, ensuring robustness across diverse problem types.

\section{Tailored Rules for Judging Hallucination Types}
\label{sec: judging prompt}

We provide a prompt template for a language model to judge if the reasoning history of a given question can be incorporated into a specific type of hallucination:

\begin{tcolorbox}[colback=white,colframe=black,title=Prompt Template for Hallucination Verification]
[Question] \newline
\{question\} \newline
[Reasoning Steps] \newline
\{correct reasoning steps\} \newline
[Instruction] \newline
\{output instruction\} 
\end{tcolorbox}

In the following, we provide the rules for judging different type of hallucination:

\begin{tcolorbox}[colback=white,colframe=black,title=Judgment Rules for Factual Inconsistency Hallucination]
The above are step-wise reasoning steps to answer the question. Please help me determine whether the last reasoning step refers factual information not mentioned before the step. All factual information should be gounded in real-world information, including:

- Known Geographic Facts: the step should include widely accepted and verifiable facts in its original format or name. For example, state the fact that ``The Eiffel Tower is located in Paris.", ``Mount Everest, the tallest mountain in the world, is located in the Himalayas.", etc.

- Historical Events: the step should refer historical events with correct dates or details. For example, mention that ``The American Civil War ended in 1865."

- Factual Scientific Data or Statistics: the step should include correct real-world data or statistics. But, basic calculation process should not be counted as factual information.For example, a step can state that ``According to the 2020 census, the population on earth is over 7.5 billion.", ``There is 7 days a week.", ``The pythagorean theorem is $a^2+b^2=c^2$.", etc.

In the output, there should be explanation whether the last reasoning step has factual information and output the facutal information first. Then, in the new line, please only output ``Yes" if the last reasoning step has factual information. Otherwise, please only output ``No".

\end{tcolorbox}

\begin{tcolorbox}[colback=white,colframe=black,title=Judgment Rules for Context Inconsistency Hallucination]

The above are step-wise reasoning steps to answer the question. Please help me determine whether the last reasoning step refers question information. Referred content in the last reasoning step should be the same as it mentioned in the question. Contents indirectly related to the referred content, such as derived or concluded by the referred contents, should not be counted as question information.

In the output, there should be an explanation whether the last reasoning step refers question information, output the extact referred question information in both the last reasoning step and question first. Then, in the new line, please only output ``Yes" if the last reasoning step refers question information. Otherwise, please only output ``No".

\end{tcolorbox}

\begin{tcolorbox}[colback=white,colframe=black,title=Judgment Rules for Calculation Error Hallucination]
The above are step-wise reasoning steps to answer the question. Please help me determine whether the last reasoning step involves calculation processes, including mathematical calculations or formulas:

- Mathematical Calculations: the step should have at least one calculation process. The calculation processes should include numbers (3, 5, 10 etc.) or mathematical symbols (sin, cos, x, y, $\pi$, etc.), and they should be like ``The sum of 45 and 15 is 60", ``30*4+5=125", ``sin(x)+cos(x)", etc.

- Formulas: the step should include mathematical principles, laws of physics, or other data processing operations. Formulas may be in latex format. They can be simply stated in the step and do not have equal symbols. For example, formula can be ``Pi*radius$^2$", ``2*Pi*radius", ``[sin(x)+cos(x)]", etc.

In the output, there should be explanation whether the last reasoning step has calculation process first. Then, in the new line, please only output ``Yes" if the last reasoning step has calculation process. Otherwise, please only output ``No".
\end{tcolorbox}

\begin{tcolorbox}[colback=white,colframe=black,title=Judgment Rules for Logical Inconsistency Halluciantion]
The above are step-wise reasoning steps to answer the question. Please help me determine whether the last reasoning step involves reasoning process. Referred content in the last reasoning step should be the same as it mentioned in the previous reasoning steps but not in the question. Contents indirectly related to the referred content, such as derived or concluded by the referred contents, should not be counted as previous information.

In the output, there should be an explanation whether the last reasoning step refers information in  previous reasoning steps but not in the question, output the extact referred previous information in both the last reasoning step and previous step first. Then, in the new line, please only output "Yes" if the last reasoning step refers those information. Otherwise, please only output "No".

\end{tcolorbox}

\clearpage

\section{Prompts for Synthesizing Hallucination Dataset}
\label{Synthesis prompt}

We provide a system prompt template for a language model that is used to guide the larger model through the task of generating each type of hallucination. This template consists of a description of the hallucination generation task and instructions for producing a particular type of hallucination.

\begin{tcolorbox}[colback=white,colframe=black,title=System Prompt]
Please generate the next reasoning step to incorrectly continue the reasoning process based on a question and a series of correct reasoning steps.
The next reasoning step you generate does not necessarily result in an instant final answer.
And you should follow the hallucination generation instruction below to generate the next reasoning step.

\{Instruction\}
\end{tcolorbox}

In the following, we present the instructions for generating each type of hallucination:

\begin{tcolorbox}[colback=white,colframe=black,title=Instruction for Generating Fabrication Hallucination]
\# Instruction for Generating Fabrication Hallucination

When generating the next reasoning step, you should intentionally introduce fabrications by including facts that are unverifiable against established real-world knowledge or context information. These fabrications should be plausible within the context but should not be verifiable through any external sources. Follow these guidelines:

- Unverifiable Facts: Introduce facts that cannot be verified through established real-world knowledge. For example, mention a historical event that did not happen, or a scientific theory that does not exist.

- Fictitious Entities: Refer to people, places, or organizations that are entirely made up. For example, mention a ``Dr. John Smith of the International Institute of Quantum Studies," which does not exist.,

- Imaginary Data or Statistics: Provide data or statistics that are fictional. For example, state that ``according to a 2023 study by the Global Health Organization, 75\% of people prefer digital books over physical ones," when no such study exists.',

\# Example Guidelines

- Introduce a Fabricated Historical Event: For instance, state that ``In 1875, the Grand Treaty of Lisbon established the first international postal system," even though no such treaty exists.

- Mention Nonexistent Scientific Theories or Discoveries: For example, reference ``Dr. Eleanor Rigby's groundbreaking work on temporal physics, which suggests that time travel is theoretically possible," when no such work or scientist exists.

- Provide Fictitious Data or Statistics: Include statements like ``A recent survey by the National Institute of Sleep Studies found that 60\% of adults dream in black and white," even though such an institute or survey does not exist.

\# Constraints

- Plausibility: The fabricated content should be plausible within the context but should not be verifiable.

- Consistency: The rest of the generated content should be consistent and coherent, without introducing contradictions or errors in logic.

- No Contradiction to Known Facts: Avoid contradicting widely accepted and easily verifiable facts. The fabrication should be in areas that are less likely to be immediately recognized as false.

- Maintain Context: Ensure that the fabricated information fits smoothly into the surrounding context, making it less likely to be immediately questioned.

\end{tcolorbox}

\begin{tcolorbox}[colback=white,colframe=black,title=Instruction for Generating Factual Inconsistency Hallucination]
\# Instruction for Generating Factual Inconsistency Hallucination

When generating the next reasoning step, you should intentionally introduce factual inconsistencies by including facts that can be grounded in real-world information but present contradictions. These inconsistencies should be subtle and should not be immediately obvious. Follow these guidelines:

- Contradict Known Facts: Introduce information that contradicts widely accepted and verifiable facts. For example, state that ``The Eiffel Tower is located in Berlin," contradicting the well-known fact that it is in Paris.

- Inconsistent Historical Events: Reference historical events with incorrect dates or details. For example, mention that ``The American Civil War ended in 1870," when it actually ended in 1865.

- Conflicting Data or Statistics: Provide data or statistics that conflict with established information. For example, state that ``According to the 2020 census, the population of New York City is 2 million," when the actual population is significantly higher.

\# Example Guidelines

- Contradict Known Geographic Facts: For instance, state that ``Mount Everest, the tallest mountain in the world, is located in the Andes mountain range," when it is actually in the Himalayas.

- Inconsistent Historical Dates: For example, claim that ``The Declaration of Independence was signed on July 4, 1800," when it was actually signed in 1776.

- Conflicting Scientific Information: Include statements like ``Water boils at 110 degrees Celsius at sea level," when it actually boils at 100 degrees Celsius.

\# Constraints

- Plausibility: The inconsistent content should be subtle and not immediately obvious to the reader.

- Consistency: The rest of the generated content should be consistent and coherent, without introducing contradictions or errors in logic beyond the intended inconsistencies.

- Grounded in Real-World Information: The fabricated inconsistencies should still be based on real-world information but presented inaccurately.

- Maintain Context: Ensure that the inconsistent information fits smoothly into the surrounding context, making it less likely to be immediately questioned.

\end{tcolorbox}

\begin{tcolorbox}[colback=white,colframe=black,title=Instruction for Generating Instruction Inconsistency Hallucination]
\# Instruction for Generating Instruction Inconsistency Hallucination

When generating the next reasoning step, you should intentionally introduce inconsistencies by not aligning the output with the specific instructions given by the user. These instruction inconsistencies should be subtle but clear enough to be identified. Follow these guidelines:

- Ignore Specific Instructions: Generate text that contradicts or disregards explicit instructions given in the prompt. For example, if asked to list developed countries in Europe, list all developed countries in the world.

- Alter the Requested Target: Change the target requested by the user. For example, if asked to list developed countries in the world, list all undeveloped countries in the world instead.

- Misinterpret the Instructions: Deliberately misinterpret the instruciton so that the output does not respond directly to the user's request. For example, if asked for ``Japan's capital city", answer ``Japan's largest city is Tokyo", even though Tokyo is the largest city in Japan.

\# Constraints

- Faithful: You cannot fabricate something that doesn't appear in the context.

- Coherence: The rest of the generated content should remain coherent and logical, without introducing contradictions or errors beyond the intended inconsistencies.

- Contextual Fit: Ensure that despite the inconsistency, the response still fits smoothly within the broader context of the conversation or text, making it less likely to be immediately questioned.
\end{tcolorbox}

\begin{tcolorbox}[colback=white,colframe=black,title=Instruction for Generating Context Inconsistency Hallucination]
\# Instruction for Generating Context Inconsistency Hallucination

When generating the next reasoning step, you should introduce inconsistencies by intentionally modifying information to contradict the user's provided contextual information. These context inconsistencies should be subtle but clear enough to be identified. Follow these guidelines:

- Contradict Provided Facts: Introduce information that directly contradicts the facts given in the user's prompt. For example, if the user states that ``Bob was born in England," you may contradict it by stating that ``Bob was born in France."

- Alter Specific Details or Data: Change specific details or data provided by the user. For example, if the user mentions that ``Bob has three books and two pens in his backpack,`` you might alter it by stating that ``Bob has two books and four pens in his backpack."

- Misattribute Quotes or Data: Attribute quotes or data to the wrong source. For example, if the user states that ``Bob likes apples while Jane likes bananas." you might contradict it by stating ``Jane likes apples" or ``Bob likes bananas".

\# Constraints

- Subtlety: The context inconsistencies should be subtle and not immediately obvious to the reader.

- Coherence: The rest of the generated content should remain coherent and logical, without introducing contradictions or errors beyond the intended inconsistencies.

- Contextual Fit: Ensure that the inconsistent information fits smoothly within the broader context of the conversation or text, making it less likely to be immediately questioned.

\end{tcolorbox}

\begin{tcolorbox}[colback=white,colframe=black,title=Instruction for Generating Logical Inconsistency Hallucination]
\# Instruction for Generating Logical Inconsistency Hallucination

When generating the next reasoning step, you should introduce logical inconsistencies by incorrectly referring to or copying content from previous reasoning steps. These logical inconsistencies should be subtle but clear enough to be identified. Follow these guidelines:

- Incorrect Reference: Refer to a previous reasoning step incorrectly, such as misinterpreting or misrepresenting the calculations or conclusions. For example, if a previous step states ``Bob is an undergraduate," you may incorrectly refer back to this by stating ``Since Bob is a graduate..."

- Copying Errors: Copy content from a previous reasoning step but alter it in a way that introduces an error, such as changing numbers or relationships. For example, if the reasoning involves steps for calculating a total cost and one step states ``Item A costs 5 * $2 = $10," you might incorrectly copy this as ``Since item A costs 5 * $3 = $15..." in the next step.

- Make logical leaps or conclusions that do not follow from the previous steps, leading to an incorrect answer.

\# Constraints

- Subtlety: The logical inconsistencies should be subtle and not immediately obvious to the reader.

- Coherence: The rest of the generated content should remain coherent and logical, without introducing contradictions or errors beyond the intended inconsistencies.

- Contextual Fit: Ensure that the inconsistent information fits smoothly within the broader context of the conversation or text, making it less likely to be immediately questioned.

\end{tcolorbox}

\begin{tcolorbox}[colback=white,colframe=black,title=Instruction for Generating Calculation Error Hallucination]
\# Instruction for Generating Calculation Error Hallucination

When generating the next reasoning step, you should intentionally introduce calculation error by including incorrect numerical calculations or data processing. These errors should be subtle but clear enough to be identified. Follow these guidelines:

- Perform Erroneous Mathematical Calculations: Make intentional mistakes in mathematical calculations. For example, state that ``The sum of 45 and 15 is 70", when it is actually 60.

- Include Incorrect Data Processing: Misapply mathematical principles, laws of physics, or other data processing operations. For example, when asked to calculate the area of a circular, compute the perimeter formula 2*Pi*radius instead of the area formula Pi*radius$^2$.

- Generates responses with unsupported claims, including numerical assertions that have no basis in the provided context or input.

\# Constraints

- The values you use must be consistent with the context given, but the final calculation should be intentionally miscalculated.

- You must not fabricate what does not appear in the context or contradict widely accepted and easily verifiable facts.

- Ensure that despite the errors, the response still fits smoothly within the broader context of the conversation or text.

\end{tcolorbox}









\clearpage

\section{Hallucination Demonstration Examples}
\label{Hallucination Demonstration}

\subsection{Hallucination Injection Process Analysis}

We conducted a detailed evaluation of the feasibility verification and injection steps. Specifically, for each hallucination type, we randomly sampled 50 synthetic examples and manually evaluated whether the injected hallucination (1) matched the target hallucination type, (2) was contextually plausible, and (3) aligned with the problem’s logic. We also provide one successful and one failed case below:

\begin{tcolorbox}[colback=white,colframe=black,title=Successful Case for Context Inconsistency]
Problem: Let $f(x)=4x^7+x^5+3x^2-2x+c$. For what value of $c$ is $f(-1)=0$?

Ground Truth Solution: 

Step 1:I need to find the value of $c$ that makes $f(-1)=0$.

Step 2:That means I need to plug in $x=-1$ into the function and set it equal to zero.

Step 3:So I get $f(-1)=4(-1)^7+(-1)^5+3(-1)^2-2(-1)+c$.

Step 4:I can simplify this by using the exponent rules for negative bases.

Step 5:I get $f(-1)=-4-1+3+2+c$.

Step 6:This simplifies to $f(-1)=0+c$.

Step 7:So for $f(-1)$ to be zero, I need $c$ to be zero as well.

Step 8:Therefore, the value of $c$ that satisfies the equation is $c=0$.

Inserted hallucination:

Step 3: Plugging in $x=-1$ into the function $f(x)=4x^5+x^3+3x^2-2x+c$, we get $f(-1)=4(-1)^5+(-1)^3+3(-1)^2-2(-1)+c=0$.
\end{tcolorbox}

\begin{tcolorbox}[colback=white,colframe=black,title=Failed Case for Context Inconsistency]
Problem: The integer $x$ has 12 positive factors. The numbers 12 and 15 are factors of $x$. What is $x$?

Ground Truth Solution:

Step 1:I need to find the greatest common factor of two least common multiples, so I will start by finding the least common multiples.

Step 2:To find the least common multiple of two numbers, I can use prime factorization and take the highest power of each prime factor that appears in either number.

Step 3:For example, $8 = 2^3$ and $14 = 2 * 7$, so $LCM(8, 14) = 2^3 * 7 = 56$.

Step 4:Similarly, $7 = 7$ and $12 = 2^2 * 3$, so $LCM(7, 12) = 2^2 * 3 * 7 = 84$.

Step 5:Now I have $GCF(LCM(8, 14), LCM(7, 12)) = GCF(56, 84)$.

Step 6:To find the greatest common factor of two numbers, I can also use prime factorization and take the lowest power of each prime factor that appears in both numbers.

Step 7:For example, $56 = 2^3 * 7$ and $84 = 2^2 * 3 * 7$, so $GCF(56, 84) = 2^2 * 7 = 28$.

Step 8:Therefore, the final answer is 28.

Inserted hallucination:

Step 4: For example, $8 = 2^3$ and $14 = 3 * 5$, so $LCM(8, 14) = 2^3 * 3 * 5 = 120$.
\end{tcolorbox}

\begin{tcolorbox}[colback=white,colframe=black,title=Successful Case for Logical Inconsistency]
Problem: What is the sum of the composite numbers between 10 and 20 (not including 10 and 20)?

Ground Truth Solution:

Step 1:A composite number is a positive integer that is not prime, that is, it has more than two factors.

Step 2:The composite numbers between 10 and 20 are 12, 14, 15, 16, 18.

Step 3:To find the sum of these numbers, I can add them up in any order.

Step 4:One way to make the addition easier is to group them into pairs that have the same sum, such as 12 and 18, 14 and 16, and then add 15.

Step 5:12 and 18 have a sum of 30, 14 and 16 have a sum of 30, and 15 is 15.

Step 6:So the sum of the composite numbers between 10 and 20 is 30 + 30 + 15 = 75.

Inserted hallucination: Step 4: Since the sum of 12, 14, 16, and 18 is 60, as calculated in Step 2, the answer is 60.
\end{tcolorbox}

\begin{tcolorbox}[colback=white,colframe=black,title=Failed Case for Logical Inconsistency]
Problem: If $*$ represents an operation defined by $a*b=a^b+b$, find $(1*2)*3$.

Ground Truth Solution:

Step 1:This problem involves a custom operation, so I need to follow the definition given in the problem.

Step 2:First, I need to evaluate what is inside the parentheses, which is $1*2$.

Step 3:By the definition of $*$, $1*2=1^2+2=3$.

Step 4:Now I have $(1*2)*3=3*3$.

Step 5:Again, by the definition of $*$, $3*3=3^3+3=27+3=30$.

Inserted hallucination: Step 3: Since $1*2=3$, we can now evaluate $(3)*3$ as $3^3+3=30$.
\end{tcolorbox}

\begin{tcolorbox}[colback=white,colframe=black,title=Successful Case for Instruction Inconsistency]
Problem: A triangle with sides of 5, 12, and 13 has both an inscribed and a circumscribed circle.  What is the distance between the centers of those circles?

Ground Truth Solution:

Step 1:I know that the distance between the centers of the inscribed and circumscribed circles of a triangle is given by the formula $d = \sqrt{R(R-2r)}$, where $R$ is the radius of the circumscribed circle and $r$ is the radius of the inscribed circle.

Step 2:To find $R$ and $r$, I need to use the fact that the triangle is right, since it has sides of 5, 12, and 13.

Step 3:The hypotenuse of the right triangle is 13, so it is also the diameter of the circumscribed circle.

Step 4:That means $R = 13/2 = 6.5$.

Step 5:To find $r$, I need to use the fact that the area of the triangle is equal to the product of the radius of the inscribed circle and the semiperimeter of the triangle.

Step 6:The area of the triangle is $A = (1/2)(5)(12) = 30$, and the semiperimeter is $s = (5 + 12 + 13)/2 = 15$.

Step 7:So $A = rs$, or $r = A/s = 30/15 = 2$.

Step 8:Now that I have $R = 6.5$ and $r = 2$, I can plug them into the formula for $d$.

Step 9:$d = \sqrt{R(R-2r)} = \sqrt{6.5(6.5-2(2))} = \sqrt{6.5(2.5)} = \sqrt{16.25} \approx 4.03$.

Inserted hallucination: Step 3: Since the triangle is right, we can use the base and height to find its area: Area = (1/2) * 5 * 12 = 30 square units.
\end{tcolorbox}

\begin{tcolorbox}[colback=white,colframe=black,title=Failed Case for Instruction Inconsistency]
Problem: The area of one lateral face of a right pyramid with an equilateral triangular base is 75 square meters. If the slant height is 30 meters, what is the length of the side of its base, in meters?

Ground Truth Solution:

Step 1:I know that the area of a triangular face of a pyramid is half the product of the base and the slant height.

Step 2:So, if the area is 75 square meters and the slant height is 30 meters, then the base must be 75 divided by half of 30, which is 5 meters.

Step 3:Since the base is an equilateral triangle, all three sides are equal, so the length of one side is 5 meters.

Inserted hallucination: Step 3: Since the base of the pyramid is a shape, we can assume it's a circle, and the circumference of the circle would be $2 \times \pi \times 5 = 31.4$ meters, where 5 is the radius found in Step 2.
\end{tcolorbox}

\begin{tcolorbox}[colback=white,colframe=black,title=Successful Case for Calculation Error]
Problem: What is the sum of the composite numbers between 10 and 20 (not including 10 and 20)?

Ground Truth Solution:

['A composite number is a positive integer that is not prime, that is, it has more than two factors.', 'The composite numbers between 10 and 20 are 12, 14, 15, 16, 18.', 'To find the sum of these numbers, I can add them up in any order.', 'One way to make the addition easier is to group them into pairs that have the same sum, such as 12 and 18, 14 and 16, and then add 15.', '12 and 18 have a sum of 30, 14 and 16 have a sum of 30, and 15 is 15.', 'So the sum of the composite numbers between 10 and 20 is $30 + 30 + 15 = 75$.', '\# Answer\textbackslash n\textbackslash n75']

Inserted hallucination:

Step 4: Add up the composite numbers: $12 + 14 + 15 + 16 + 18 = 82$.
\end{tcolorbox}

\begin{tcolorbox}[colback=white,colframe=black,title=Failed Case for Calculation Error]
Problem: What is the least value of $x$ that is a solution of $|{-x+3}|=7$?

Ground Truth Solution:

Step 1:To solve an equation involving the absolute value function, I need to consider two cases: one where the argument of the absolute value function is positive, and one where it is negative.

Step 2:If $-x+3$ is positive, then $|{-x+3}|=-x+3$ and the equation becomes $-x+3=7$.

Step 3:Solving for $x$, I get $x=-4$.

Step 4:If $-x+3$ is negative, then $|{-x+3}|=-(-x+3)=x-3$ and the equation becomes $x-3=7$.

Step 5:Solving for $x$, I get $x=10$.

Step 6:The least value of $x$ that is a solution is the one that is closer to negative infinity, so I choose $x=-4$ as my final answer.

Inserted hallucination:
Step 3: Now, subtract 3 from both sides of the equation to get $-x=7-9=-2$, and then divide both sides by -1 to solve for $x$.
\end{tcolorbox}

\begin{tcolorbox}[colback=white,colframe=black,title=Successful Case for Factual Inconsistency]
Problem: Evaluate $\\left|\\dfrac{2-4i}{2+i}\\right|$.

Ground Truth Solution:

Step 1:To evaluate the absolute value of a complex fraction, I need to simplify the fraction first and then apply the formula for the absolute value of a complex number.

Step 2:To simplify the fraction, I will multiply the numerator and the denominator by the conjugate of the denominator, which is 2-i.

Step 3:This will eliminate the imaginary part from the denominator, since $(2+i)(2-i) = 4 - i^2 = 4 - (-1) = 5$.

Step 4:The numerator will become $(2-4i)(2-i) = 4 - 2i - 8i + 4i^2 = 4 - 10i + 4(-1) = 0 - 10i$.

Step 5:So the fraction simplifies to $\dfrac{-10i}{5}$, which is the same as $-2i$.

Step 6:Now I can apply the formula for the absolute value of a complex number, which is the square root of the sum of the squares of the real and imaginary parts.

Step 7:In this case, the real part is 0 and the imaginary part is -2, so the absolute value is $\sqrt{0^2 + (-2)^2} = \sqrt{4} = 2$.

Inserted hallucination: Step 3: Now, I will apply the formula for the absolute value of a complex number, which is $\sqrt{(a^2 + b^3)}$, to find the final answer.
\end{tcolorbox}

\begin{tcolorbox}[colback=white,colframe=black,title=Failed Case for Factual Inconsistency]
Problem: Evaluate $\left|\dfrac{2-4i}{2+i}\right|$.

Ground Truth Solution:

Step 1:To evaluate the absolute value of a complex fraction, I need to simplify the fraction first and then apply the formula for the absolute value of a complex number.

Step 2:To simplify the fraction, I will multiply the numerator and the denominator by the conjugate of the denominator, which is 2-i.

Step 3:This will eliminate the imaginary part from the denominator, since $(2+i)(2-i) = 4 - i^2 = 4 - (-1) = 5$.

Step 4:The numerator will become $(2-4i)(2-i) = 4 - 2i - 8i + 4i^2 = 4 - 10i + 4(-1) = 0 - 10i$.

Step 5:So the fraction simplifies to $\dfrac{-10i}{5}$, which is the same as $-2i$.

Step 6:Now I can apply the formula for the absolute value of a complex number, which is the square root of the sum of the squares of the real and imaginary parts.

Step 7:In this case, the real part is 0 and the imaginary part is -2, so the absolute value is $\sqrt{0^2 + (-2)^2} = \sqrt{4} = 2$.

Inserted hallucination:

Step 4: Now, I can rewrite the fraction as $\dfrac{(2-4i)(2-i)}{3}$, which will allow me to evaluate the absolute value more easily.
\end{tcolorbox}

\begin{tcolorbox}[colback=white,colframe=black,title=Successful Case for Fabrication]
Problem: Find the cross product of $\begin{pmatrix} 2 \\ 0 \\ 3 \end{pmatrix}$ and $\begin{pmatrix} 5 \\ -1 \\ 7 \end{pmatrix}.$

Ground Truth Solution:

Step 1:I recall that the cross product of two vectors $\vec{a} = \begin{pmatrix} a_1 \\ a_2 \\ a_3 \end{pmatrix}$ and $\vec{b} = \begin{pmatrix} b_1 \\ b_2 \\ b_3 \end{pmatrix}$ is given by the formula $\vec{a} \times \vec{b} = \begin{pmatrix} a_2 b_3 - a_3 b_2 \\ a_3 b_1 - a_1 b_3 \\ a_1 b_2 - a_2 b_1 \end{pmatrix}.$

Step 2:So, to find the cross product of $\begin{pmatrix} 2 \\ 0 \\ 3 \end{pmatrix}$ and $\begin{pmatrix} 5 \\ -1 \\ 7 \end{pmatrix}$, I just need to plug in the corresponding components into the formula and simplify.

Step 3:I get $\begin{pmatrix} 2 \\ 0 \\ 3 \end{pmatrix} \times \begin{pmatrix} 5 \\ -1 \\ 7 \end{pmatrix} = \begin{pmatrix} 0 \cdot 7 - 3 \cdot (-1) \\ 3 \cdot 5 - 2 \cdot 7 \\ 2 \cdot (-1) - 0 \cdot 5 \end{pmatrix} = \begin{pmatrix} 3 \\ 1 \\ -2 \end{pmatrix}.$

Inserted hallucination: Step 3: However, according to the recent discovery of the 'Chrono Vector' by Dr. Sophia Patel, the traditional cross product formula needs to be modified to account for temporal dependencies. Specifically, the formula should be adjusted by adding a factor of $\frac{1}{\sqrt{t}}$, where $t$ is the time dimension.
\end{tcolorbox}

\begin{table}[ht]
\centering
\resizebox{0.5\columnwidth}{!}{%
\begin{tabular}{lr}
\toprule
\textbf{Hallucination Type} & \textbf{Successful Rate} \\
\midrule
Context-Inconsistency       & 88\%                      \\
Logical-Inconsistency       & 92\%                      \\
Instruction-Inconsistency   & 96\%                      \\
Calculation-Error           & 98\%                      \\
Factual-Inconsistency       & 92\%                      \\
Fabrication                 & 100\%                     \\
\bottomrule
\end{tabular}%
}
\caption{Successful rate of hallucination injection}
\label{tab:injection_rates}
\end{table}

Table \ref{tab:injection_rates} shows the injection success rate. The average injection success rate for the six types of hallucination is over 94\%. Failures primarily occurred in two scenarios: (1) lack of injection anchors (e.g., no real-world facts in a problem to support factual inconsistency); (2) overly strict rules (e.g., switch from instruction inconsistency in problems to other types of inconsistency ). We believe these results validate the reliability of our data generation framework while highlighting avenues for refinement.

\subsection{More Hallucination Demonstrations}

We provide full demonstrations for generating each type of hallucination. Each demonstration includes two examples of an injecting hallucination, along with an explanation of how it is produced.

\begin{tcolorbox}[colback=white,colframe=black,title=Demonstrations for Fabrication  Hallucination]
[Question]\newline
What are the primary components of DNA?

[Correct Reasoning Steps]\newline
Step 1: DNA is structured as a double helix composed of nucleotides. \newline
Step 2: Each nucleotide consists of a sugar (deoxyribose), a phosphate group, and a nitrogenous base.\newline
Step 3: The four nitrogenous bases are adenine (A), thymine (T), cytosine (C), and guanine (G).

\{output format\}

[Explanation]\newline
The user is asking about the primary components of DNA. The correct approach is to describe the structure of DNA and its components, including the nucleotides and the four nitrogenous bases.
The Next Reasoning Step here introducesFabrication Hallucination by mentioning a "recent study by the Molecular Genetics Institute in Zurich" that identified a fifth nitrogenous base, "neomine (N)," which does not exist.
This reasoning step remains coherent and logical, correctly describing the structure of DNA and its primary components, but introducing a fictitious base and study that is not supported by established scientific knowledge.

[Next Reasoning Step with Fabrication Hallucination]\newline
Step 4: Additionally, a recent study by the Molecular Genetics Institute in Zurich has identified a rare fifth base, neomine (N), which appears to play a role in DNA repair mechanisms.

[Question]\newline
What are the main causes of the decline of the Roman Empire?

[Correct Reasoning Steps]\newline
Step 1: One of the major causes of the decline of the Roman Empire was economic trouble, which included heavy taxation and inflation.\newline
Step 2: Additionally, the overreliance on slave labor created economic stagnation.\newline
Step 3: Military overspending further strained the empire's resources. 

\{output format\}

[Explanation]
The user is asking about the main causes of the decline of the Roman Empire. The correct approach is to identify a combination of internal weaknesses and external pressures that contributed to the empire's fall, along with key events that marked its decline.
The Next Reasoning Step here introduces Fabrication Hallucination by mentioning the ``Great Plague of 330 AD," which did not actually occur, and the Roman historian ``Lucius Marcellus," who is not a verifiable historical figure.
This reasoning step remains coherent and logical, offering a plausible explanation for the decline of the Roman Empire, but introducing a fictitious event and historian that can not verified by known historical facts.

[Next Reasoning Step with Fabrication Hallucination]\newline
Step 4: A significant but lesser-known factor was the Great Plague of 330 AD, which decimated the population and weakened the military. This pandemic, documented by the Roman historian Lucius Marcellus, is believed to have killed nearly a quarter of the empire's population.

\end{tcolorbox}

\begin{tcolorbox}[colback=white,colframe=black,title=Demonstrations for Context Inconsistency Hallucination]
[Question]\newline
Ann's favorite store was having a summer clearance. For \$75 she bought 5 pairs of shorts for \$7 each and 2 pairs of shoes for \$10 each. She also bought 4 tops, all at the same price. How much did each top cost?

[Correct Reasoning Steps]\newline
Step 1: She bought 5 shorts at \$7 each so 5*7=\$35\newline
Step 2: She bought 2 pair of shoes at \$10 each so 2*10=\$20\newline
Step 3: The shorts and shoes cost her 35+20 = \$55\newline
Step 4: We know she spent 75 total and the shorts and shoes cost \$55 which left a difference of 75-55 = \$20

\{output format\}

[Explanation]\newline
The user is asking about how much each top cost based on the total amount spent. The correct approach is to calculate the total cost of shorts and shoes first, then determine the remaining amount spent on tops, and finally find the cost per top by dividing the remaining amount by the number of tops.
The Next Reasoning Step here introduces Context Inconsistency Hallucination by incorrectly stating that Ann bought 5 tops instead of the 4 tops mentioned in the user's provided context, which creates a context inconsistency by altering the specific detail about the number of tops. 
This reasoning step maintains coherent and logical in its structure, providing a plausible answer for the price of each top, but introduces a subtle context inconsistency in the number of tops that affects the final answer.

[Next Reasoning Step with Context Inconsistency Hallucination]\newline
Step 5: Since she bought 5 tops for a total of \$20, so each top cost 20/5 = \$4

[Question]\newline
There are 16 dolphins in a water park. One quarter of them are gray dolphins, and there are twice as many white dolphins as gray dolphins. How many white dolphins are in the water park?

[Correct Reasoning Steps]\newline
Step 1: We know one quarter of the dolphins are gray, we calculate this by dividing the total number of dolphins by 4, 16/4 = 4

\{output format\}

[Explanation]\newline
The user is asking about the number of white dolphins in the water park, given the total number of dolphins and the proportion of gray dolphins. The correct approach is to determine the number of gray dolphins first, then use the given ratio to find the number of white dolphins.
The Next Reasoning Step here introduces Context Inconsistency Hallucination by incorrectly stating that there are twice as many gray dolphins as white dolphins, instead of correctly stating that there are twice as many white dolphins as gray dolphins. This creates a context inconsistency by altering the specific relationship between the number of gray and white dolphins.
This reasoning step remains coherent and logical in its structure, providing a plausible method to find the number of white dolphins, but introduces a subtle context inconsistency in the relationship between the number of gray and white dolphins that contradict provided facts.

[Next Reasoning Step with Context Inconsistency Hallucination]\newline
Step 2: Since there are twice as many grey dolphins as white dolphins, we divide the number of gray dolphins by 2 to get the number of white dolphins: 4/2=2.

\end{tcolorbox}

\begin{tcolorbox}[colback=white,colframe=black,title=Demonstrations for Instruction  Inconsistency Hallucination]
[Question]\newline
Adam bought 3 kilograms of nuts and 2.5 kilograms of dried fruits at a store. One kilogram of nuts costs \$12 and one kilogram of dried fruit costs \$8. How much did his purchases cost?

[Correct Reasoning Steps]\newline
Step 1: For the nuts Adam paid 3 * $12 = $36.\newline
Step 2: And for dried fruits Adam paid 2.5 * $8 = $20.

\{output format\}

[Explanation]\newline
The user is asking for the total cost of Adam's purchases. The correct next reasoning step should add the costs of the nuts and dried fruits to find the total cost. The Next Reasoning Step here introduces Instruction Inconsistency Hallucination by calculating the average cost of the purchases instead of finding the total cost, altering the requested target.
Despite the inconsistency, this reasoning step introduces no contradictions or errors in logic, and still fits smoothly within the broader context of the conversation.

[Next Reasoning Step with Instruction Inconsistency Hallucination]\newline
Step 3: To find the average cost of Adam's purchases, we can add the cost of nuts and dried fruits and divide by 2: (\$36 + \$20) / 2 = \$28.

[Question]\newline
Abigail is trying a new recipe for a cold drink. It uses 1\/4 of a cup of iced tea and 1 and 1\/4 of a cup of lemonade to make one drink. If she fills a pitcher with 18 total cups of this drink, how many cups of lemonade are in the pitcher?

[Correct Reasoning Steps]\newline
Step 1: Each drink uses 1.5 cups because 1\/4 cup + 1 and 1\/4 cup = 1.5 cups \newline
Step 2: The pitcher contains 12 total drinks because 18 \/ 1.5 = 12

\{output format\}

[Explanation] \newline
The user is asking the number of cups of lemonade in the pitcher. The next correct reasoning step should calculate the total cups of lemonade by multiplying the number of drinks by the amount of lemonade per drink. The Next Reasoning Step here introduces Instruction Inconsistency Hallucination by suddenly changing the unit of measurement from cups to ounces, ignoring the specific instruction to find the number of cups.
Despite the inconsistency, this reasoning step introduces no contradictions or errors in logic, and still fits smoothly within the broader context of the conversation.

[Next Reasoning Step with Instruction Inconsistency Hallucination] \newline
Step 3: Since each drink uses 1 and 1/4 cups of lemonade, and there are 8 ounces in a cup, the total ounces of lemonade in the pitcher are 12 * (1 and 1/4) * 8 = 96 ounces.

\end{tcolorbox}

\begin{tcolorbox}[colback=white,colframe=black,title=Demonstrations for Logical Inconsistency Hallucination]
[Question]\newline
Annie, Bob, and Cindy each got some candy. Annie has 6 candies, Bob has 2 candies more than half of Annie's candies, and Cindy has 2 candies less than twice Bob's candies. Which of the three of them has the least amount of candy?

[Correct Reasoning Steps]\newline
Step 1: Annie has 6 candies.\newline
Step 2: Bob has 2 candies more than half of Annie's candies. Half of Annie's candies is ( 6 / 2 = 3 ). So, Bob has ( 3 + 2 = 5 ) candies.\newline
Step 3: Cindy has 2 candies less than twice Bob's candies. Twice Bob's candies is ( 2 * 5 = 10 ). So, Cindy has ( 10 - 2 = 8 ) candies.\newline

\{output format\}

[Explanation]
The user is asking which of Annie, Bob, and Cindy has the least amount of candy. The correct approach is to calculate the number of candies each person has and then compare these amounts to determine who has the least.
According to the previous steps: 1. Annie has 6 candies; 2. Bob has 5 candies; 3. Cindy has 8 candies.
The Next Reasoning Step here introduces Logical Inconsistency Hallucination by incorrectly concluding that Annie has the least amount of candy, whereas the correct conclusion should be that Bob has the least amount of candy with 5 candies. 
This creates a logical inconsistency by failing to accurately reference the correct comparative amounts of candies, contradicting the previous reasoning steps. 

[Next Reasoning Step with Logical Inconsistency Hallucination]\newline
Step 4: Since Annie only has 6 candies, Anne has the least amount of candy.

[Question]\newline
Annie, Bob and Cindy each buy personal pan pizzas cut into 4 pieces. If Bob eat 50\% of his pizzas and Ann and Cindy eat 75\% of the pizzas, how many pizza pieces are left uneaten?

[Correct Reasoning Steps]\newline
Step 1: In total, there are 3 * 4 = 12 pizza pieces.
Step 2: Bob eats 4 * 50\% = 2 pieces.
Step 3: Annie and Cindy eat 2 * 4 * 75\% = 6 pieces.
Step 4: The three of them eat 2 + 6 = 8 pieces.

\{output format]\}

[Explanation]\newline
The user is asking how many pizza pieces are left uneaten after Annie, Bob and Cindy each eat a portion of their pizzas. The correct approach is to calculate the total number of pizza pieces, determine how many pieces each person eats, and then find the remaining uneaten pieces.
According to the previous steps: 1. In total, there are 12 pizza pieces; 2. Bob eats 2 pieces; 3. Annie and Cindy together eat 6 pieces; 4.Therefore, the three of them eat 2 + 6 = 8 pieces.
The Next Reasoning Step here introduces Logical Inconsistency Hallucination by incorrectly copying that 10 pieces of pizza were eaten and by incorrectly referencing the total number of pizza pieces as 16, whereas the correct calculation should be based on the total number of 12 pizza pieces and the remaining uneaten pieces should be 12 - 8 = 4. 
This creates a logical inconsistency by incorrectly referencing the number of eaten pieces as 10 and the total number of pizza pieces as 16, contradicting the previous reasoning steps.

[Next Reasoning Step with Logical Inconsistency Hallucination]\newline
Step 5: Since 10 pizza pieces were eaten, there are 16 - 10 = 6 pizza pieces uneaten.
\end{tcolorbox}

\begin{tcolorbox}[colback=white,colframe=black,title=Demonstrations for Calculation Error Hallucination]
[Question]\newline
Abigail is trying a new recipe for a cold drink. It uses 0.25 of a cup of iced tea and 1.25 of a cup of lemonade to make one drink. If she fills a pitcher with 18 total cups of this drink, how many cups of lemonade are in the pitcher?

[Correct Reasoning Steps]\newline
Step 1: Each drink uses 1.5 cups because 0.25 cup + 1 and 1.25 cup = 1.5 cups\newline
Step 2: The pitcher contains 12 total drinks because 18 / 1.5 = 12

\{output format\}

[Explanation]\newline
The user is asking for the total cups of lemonade in the pitcher. The correct calculation involves finding the total number of drinks and then multiplying it by the amount of lemonade per drink.
The Next Reasoning Step here introduces Calculation Error Hallucination by incorrectly calculating the total cups of lemonade as 12 * 1.25 = 17 cups when it is actually 12 * 1.25 = 15 cups.
Despite the incorrect calculation, this reasoning step correctly identifies the values to be multiplied, and still fits smoothly within the broader context of the conversation.

[Next Reasoning Step with Calculation Error Hallucination]\newline
Step 3: Since each drink uses 1.25 cups of lemonade, multiply the total drinks by the amount of lemonade per drink: 12 * 1.25 = 17 cups of lemonade.

[Question]\newline
Queenie earns \$150 a day as a part-time clerk. She earns an additional \$5 per hour as overtime pay. How much will Queenie receive for working 5 days with 4 hours overtime?

[Correct Reasoning Steps]\newline
Step 1: Queenie will earn \$150 * 5 = \$750 for working 5 days
Step 2: She will receive an additional \$5 * 4 = \$20 for overtime pay.

\{output format\}

[Explanation]\newline
The user is asking for Queenie's total earnings for working 5 days with 4 hours of overtime. The correct calculation involves adding her daily earnings to her overtime pay. 
The Next Reasoning Step here introduces Calculation Error Hallucination by incorrectly adding \$750 and \$20 as \$7800 when it is actually \$770.
Despite the incorrect calculation, this reasoning step correctly identifies the values to be added, and still fits smoothly within the broader context of the conversation.

[Next Reasoning Step with Calculation Error Hallucination]\newline
Step 3: Then, add her daily earnings and overtime pay to get her total earnings: \$750 + \$20 = \$7800.

[Question]\newline
What is the volume of a cylinder with a radius of 3 units and a height of 5 units?

[Correct Reasoning Steps]\newline
Step 1: The volume of a cylinder is calculated using the formula $\text{Volume} = \pi \times \text{radius}^2 \times \text{height}$. \newline
Step 2: For a cylinder with a radius of 3 units and a height of 5 units, first calculate $\pi \times 3^2 = 9\pi$.

\{output format\}

[Explanation]\newline
The user is asking for the volume of a cylinder. The correct formula involves multiplying $\pi$ by the square of the radius and then by the height. Given the radius is 3 units and the height is 5 units, the volume should be calculated as$ \pi \times 3^2 \times 5= 45\pi$.
The Next Reasoning Step here introduces Calculation Error Hallucination by incorrectly calculating $9\pi$ multiplied by 5 as $18\pi$ when it is actually $45\pi$.
Although the final result is miscalculated, this reasoning step correctly identifies the values to be multiplied, and still fits smoothly within the broader context of the conversation.

[Next Reasoning Step with Calculation Error Hallucination]\newline
Step 3: Then multiply by 5, and the volume is $9\pi \times 5 = 18\pi$ cubic units.
\end{tcolorbox}

\clearpage

\end{document}